\documentclass{article}
\usepackage{iclr2027_conference,times}

\usepackage{amsmath,amsfonts,bm}

\def\1{\bm{1}}

\DeclareMathAlphabet{\mathsfit}{\encodingdefault}{\sfdefault}{m}{sl}
\SetMathAlphabet{\mathsfit}{bold}{\encodingdefault}{\sfdefault}{bx}{n}

\newcommand{\E}{\mathbb{E}}

\newcommand{\Cov}{\mathrm{Cov}}

\usepackage{amssymb}
\usepackage{graphicx}
\usepackage{subcaption} 
\usepackage{booktabs}
\usepackage{multirow}  

\usepackage{xcolor}
\usepackage{hyperref}
\usepackage{url}
\usepackage{tikz}
\usetikzlibrary{calc}
\usepackage{graphicx}
\usepackage{svg}
\usepackage{array}
\usepackage{wrapfig}
\usepackage{arydshln}

\DeclareMathOperator{\diag}{diag}
\newcommand{\Rot}{\mathcal{R}}
\newcommand{\RotTwo}{R}

\newcommand{\Normal}{\mathcal{N}}
\definecolor{mjRevision}{HTML}{B2186B}
\definecolor{mjComment}{HTML}{007A78}
\definecolor{kjRevision}{HTML}{B85C00}
\definecolor{kjComment}{HTML}{7040A0}

\newcommand{\mjr}[1]{\textcolor{red}{#1}}

\newcommand{\ourmethod}{WorldRoPE}

\title{\ourmethod{}: Probabilistic World-Centric\\
Rotary Embedding for Multi-View Attention}
\author{
Mijeong Kim$^{1}$ \hspace{2.5em}
Kyujin Lim$^{2}$ \hspace{2.5em}
Bohyung Han$^{1,2}$\\
\\
$^{1}$ECE \& $^{2}$IPAI, Seoul National University\\
\texttt{\{mijeong.kim, kyujin1011, bhhan\}@snu.ac.kr}
}
\iclrfinalcopy

\begin{document}
\maketitle


\begin{abstract}
Rotary Positional Embedding (RoPE) is widely used to encode relative positions in Transformers.
In multi-view settings, however, patch coordinates are defined independently in each image: the same 3D point can therefore have different coordinates across views, making them insufficient for establishing cross-view correspondences.
We introduce WorldRoPE, a probabilistic world-centric RoPE for multi-view attention.
At each layer, WorldRoPE back-projects each patch token into a set of volumetric Gaussian primitives in a shared world frame.
These primitives represent multiple plausible 3D supports for a token, jointly capturing alternative depth hypotheses, depth uncertainty, and the finite spatial extent of its patch frustum.
When a key token’s support is projected into a query view, it induces a distribution over where that key may appear in the query image.
WorldRoPE integrates the rotary operator over this distribution in closed form, combining rotations at the projected primitive centers according to their mixture weights with covariance-dependent attenuation.
This makes attention less sensitive to uncertain key locations in the query view, without rasterizing or warping token features.
The formulation reduces to deterministic RoPE when the support collapses to a point.
Across layers, the predicted geometry guides cross-view attention, while the updated features refine the geometry used at the next layer.
Experiments on novel view synthesis, depth estimation, and 3D geometry reconstruction demonstrate the effectiveness of WorldRoPE for multi-view Transformers.
Our project page is available at
\url{https://cv.snu.ac.kr/research/WorldRoPE}.
\end{abstract}

\section{Introduction}
\label{sec:intro}



Robotics and embodied perception require an understanding of the 3D world from observations taken at different viewpoints.
Multi-view Transformers aggregate these observations through attention, with positional encoding introducing spatial information into token interactions.
However, standard image-space Rotary Positional Embedding (RoPE)~\citep{su2024roformer,heo2024ropevit} encodes relative displacement using coordinates defined independently within each image.
The same 3D scene point can therefore have different coordinates across views, while nearby image locations may correspond to objects far apart in 3D.
These image-space relationships alone are insufficient for establishing cross-view correspondences.

Recent camera-aware approaches incorporate viewing geometry into attention through relative camera poses and projective transformations, as in GTA~\citep{miyato2024gta} and PRoPE~\citep{li2026prope}.
More recently, ray-based rotary encodings~\citep{wu2026rayrope,xie2026urope} further locate tokens along camera rays using predicted depths or predefined depth anchors and project these locations into the query view.
An image patch, however, observes a volumetric region spanning its viewing frustum, rather than positions confined to camera rays, and may contain multiple objects at different depths.
For example, a patch spanning a window frame and the scene behind it contains distinct depth cues, while the corresponding depth estimates may remain uncertain.
In addition, the rays passing through a finite image patch span a volumetric frustum in 3D.
A model of this 3D support should therefore jointly capture alternative depth hypotheses, depth uncertainty, and the finite spatial extent of the patch frustum.

To address these challenges, we introduce \ourmethod{}, a probabilistic world-centric RoPE for multi-view Transformers.
At each layer, we predict depth hypotheses, their uncertainty, and mixture weights from token features.
Using these predictions, \ourmethod{} back-projects each patch token into a set of volumetric Gaussian primitives in the shared world space.
These primitives represent multiple plausible 3D supports for a token, jointly capturing alternative depth hypotheses, depth uncertainty, and the finite spatial extent of its patch frustum.
Projecting these primitives into the query view defines a distribution over where each key token may appear.
We integrate the rotary operator over this distribution to guide cross-view attention, and the updated features inform geometry prediction at the next layer.
Table~\ref{tab:position_encoding_comparison} summarizes the differences from existing camera encodings.

We present a principled mathematical analysis of \ourmethod{} and examine its properties in multi-view attention.
First, the rotary expectation has a closed form that combines rotations at the projected primitive centers according to their mixture weights, with covariance-dependent attenuation.
The resulting operator guides attention without explicitly rasterizing or warping token features.
Second, the mixture operator is equivalent to a single scaled rotation for each rotary frequency block.
The inverse of its effective attenuation can be interpreted as a token-adaptive temperature.
Third, the formulation reduces to deterministic RoPE when the positional distribution collapses to a point.
Finally, predicted geometry guides attention at each layer, while the updated features inform geometry prediction in the next layer.

Experiments on novel view synthesis, depth estimation, and 3D geometry reconstruction demonstrate the effectiveness of \ourmethod{} for multi-view Transformers.
Ablations and visualizations further examine the proposed geometric representation and rotary integration.
Our main contributions are summarized as follows:
\begin{itemize}
    \item We introduce \ourmethod{}, a probabilistic world-centric RoPE that models token support with volumetric Gaussian primitives, jointly capturing alternative depth hypotheses, depth uncertainty, and the finite spatial extent of the patch frustum.
    \item We derive a closed-form rotary expectation over projected Gaussian distributions and integrate it into multi-view attention, coupling geometry prediction with feature updates across layers.
    \item Extensive experiments demonstrate the effectiveness of \ourmethod{} across novel view synthesis, depth estimation, and 3D geometry reconstruction.
\end{itemize}


\section{Related Work}
\label{sec:related}

\paragraph{Positional encodings for Transformers.}
\begin{table}[t]
    \centering
    \caption{
    Comparison of camera encodings for multi-view attention.
    Our method predicts depth hypotheses and their uncertainty for each token, representing possible 3D locations with volumetric Gaussian primitives.
    $\dagger$: URoPE adopts predefined depth anchors across heads, while our method combines predicted hypotheses.
    }
    \vspace{-2mm}
    \scalebox{0.78}{
        \setlength\tabcolsep{5.5pt}
        \begin{tabular}{l|cccc|l}
            \toprule
            \textbf{Method} & \textbf{Depth} & \textbf{Uncertainty} & \textbf{Multi-Hypothesis} & \textbf{Spatial Extent} & \textbf{Token Geometry} \\
            \midrule
            Pl\"ucker Ray~\citep{jin2024lvsm} & --        & --         & --                    & --         & Camera rays                  \\
            GTA~\citep{miyato2024gta}        & --        & --         & --                    & --         & --                           \\
            PRoPE~\citep{li2026prope}        & --        & --         & --                    & --         & --                           \\
            \midrule
            RayRoPE~\citep{wu2026rayrope}    & Predicted & \checkmark & --                    & --         & Ray segments                 \\
            URoPE~\citep{xie2026urope}       & Fixed     & --         & $\checkmark^{\dagger}$ & --         & Points on rays               \\
            \midrule
            \textbf{\ourmethod{} (ours)}    & Predicted & \checkmark & \checkmark            & \checkmark & \textbf{Volumetric Gaussians} \\
            \bottomrule
        \end{tabular}
    }
    \label{tab:position_encoding_comparison}
    \vspace{-3mm}
\end{table}
Early approaches encode token positions using fixed or learned absolute embeddings~\citep{vaswani2017attention,dosovitskiy2020vit}.
Subsequent works~\citep{liu2021swin, press2021train,raffel2020exploring,shaw2018self,su2024roformer} instead encode relative positions between tokens.
Among them, RoPE~\citep{su2024roformer} encodes relative displacement by rotating queries and keys according to their coordinates, supporting generalization across positional ranges.
Recent studies adapt RoPE to multidimensional visual coordinates for image recognition~\citep{heo2024ropevit,liu2026spiral}, 3D reconstruction~\citep{zhang2024geolrm, ren2025panosplatt3r}, generation~\citep{zhuo2024lumina, li2025head, yang2025cogvideox,yesiltepe2026infinity,zhao2025riflex}, and video understanding~\citep{wei2025videorope}. 
Other studies generalize RoPE using learned commuting matrices or Lie-group parameterizations~\citep{yu2025comrope,ostmeier2024liere}.
While these methods apply positional transformations to deterministic token coordinates, \ourmethod{} integrates RoPE over probabilistic token-position distributions for multi-view attention.

\paragraph{Camera-aware positional encodings for multi-view Transformers.}
Multi-view Transformers aggregate information across images for tasks such as novel view synthesis~\citep{jin2024lvsm,sajjadi2022scene,venkat2023geometry,wang2026clift,kim2026scaling}, depth estimation, feed-forward 3D reconstruction~\citep{jang2025pow3r,keetha2026mapanything,lin2025depth,zhang2024gs,jia2025h3r}, and world modeling~\citep{hyworld2026,lingbot-world}.
To encode viewing geometry, earlier approaches concatenate camera raymaps or Pl\"ucker raymaps with image features~\citep{jin2024lvsm}.
Recent approaches integrate camera information into RoPE to encode spatial relationships across views.
GTA~\citep{miyato2024gta} and PRoPE~\citep{li2026prope} directly transform token features using camera extrinsic matrices and projection matrices, respectively.
More recently, URoPE~\citep{xie2026urope} and RayRoPE~\citep{wu2026rayrope} use ray-based geometry for rotary encoding in the query camera frame, with predefined anchors and predicted depth, respectively.
In contrast, \ourmethod{} represents the 3D support of each token as a mixture of volumetric Gaussian primitives in a shared world frame.

\section{\ourmethod{}}
\label{sec:method}

\begin{figure}[t]
    \centering
    \begin{subfigure}[b]{0.99\linewidth}
        \centering
    \includegraphics[width=\linewidth]{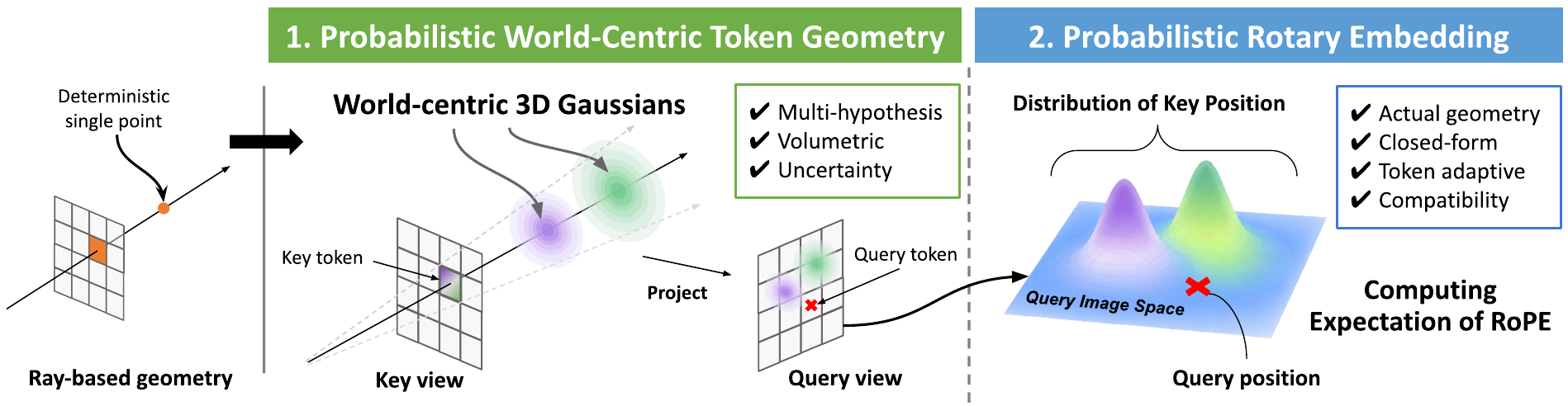}
          \vspace{-3mm}
    \end{subfigure}
    \caption{WorldRoPE overview.
    We model \textbf{probabilistic token positions} with \textbf{world-space Gaussian primitives}, capturing multiple depths, uncertainty, and patch extent.
    Their projections define possible key locations in the query view, over which \textbf{closed-form RoPE integration} guides multi-view attention without rasterizing or warping token features.}
    \label{fig:overview}
    \vspace{-2mm}
\end{figure}


We propose \ourmethod{}, a probabilistic world-centric Rotary Positional Embedding (RoPE) for multi-view attention, as shown in Figure~\ref{fig:overview}.
After reviewing notation and RoPE in Section~\ref{subsection:preliminary}, we model probabilistic token geometry with world-space Gaussian primitives in Section~\ref{subsection:GS_reconstruction}.
We then project these primitives into the query view and integrate the rotary operator over their positional distributions in Section~\ref{subsection:token_splat}.
Finally, we analyze the properties of \ourmethod{} in Section~\ref{subsection:principle_analysis}.

%


\subsection{Preliminaries}
\label{subsection:preliminary}


\paragraph{Notation.}
Consider a patch $j$ in a query view $q$ and a patch $i$ in a key view $k$.
Their coordinates are
$\mathbf{u}_{q,j}=(u_{q,j},v_{q,j})^\top$ and
$\mathbf{u}_{k,i}=(u_{k,i},v_{k,i})^\top$.
For each view $r$, let $\mathbf{K}_r$ and $(\mathbf{R}_r,\mathbf{t}_r)$ denote its intrinsic and extrinsic camera parameters.
At layer $\ell$, $\mathbf{f}_{r,s}^{(\ell)}$ denotes the feature of patch $s$ in view $r$ before the attention projections.
Suppressing the layer index for brevity, we compute
$\mathbf{q}_{q,j}=\mathbf{W}_Q\mathbf{f}_{q,j}\in\mathbb{R}^d$, $\mathbf{k}_{k,i}=\mathbf{W}_K\mathbf{f}_{k,i}\in\mathbb{R}^d$, and
$\mathbf{v}_{k,i}=\mathbf{W}_V\mathbf{f}_{k,i}\in\mathbb{R}^d$,
where $d$ is the attention-head dimension.

%
%

\paragraph{Rotary Positional Embedding (RoPE).}
Standard RoPE~\citep{su2024roformer,heo2024ropevit} encodes relative positions by rotating queries and keys before computing their inner product.
For a 2D image coordinate $\mathbf{u}=(u,v)^\top$, the $b^\text{th}$ channel pair is rotated by $\bm{\omega}_b^\top\mathbf{u}$, where $\bm{\omega}_b\in\mathbb{R}^2$ is its frequency vector.
The full operator is given by
\begin{align}
    \mathcal{R}(\mathbf{u})
    &=
    \operatorname{blockdiag}\!\left(
    R(\bm{\omega}_1^\top\mathbf{u}),
    \ldots,
    R(\bm{\omega}_{d/2}^\top\mathbf{u})
    \right), \quad \text{where} \quad
    R(\theta) =
    \begin{bmatrix}
        \cos\theta & -\sin\theta\\
        \sin\theta & \cos\theta
    \end{bmatrix}.
    \label{eq:rope}
\end{align}
Applying this operator gives the rotated query and key, denoted by $\mathcal{R}(\mathbf{u}_{q,j})\mathbf{q}_{q,j}$ and $\mathcal{R}(\mathbf{u}_{k,i})\mathbf{k}_{k,i}$, respectively.
When the coordinates are expressed in a common image frame, the inner product of the rotated query and key depends only on their relative displacement, since $\mathcal{R}(\mathbf{u}_{q,j})^\top \mathcal{R}(\mathbf{u}_{k,i}) = \mathcal{R}(\mathbf{u}_{k,i}-\mathbf{u}_{q,j})$.
In the following, we omit the frequency index $b$ when discussing a single block.

\subsection{Probabilistic World-Space Token Geometry}
\label{subsection:GS_reconstruction}

\begin{figure}[t]
    \centering
    \includegraphics[width=0.99\linewidth]{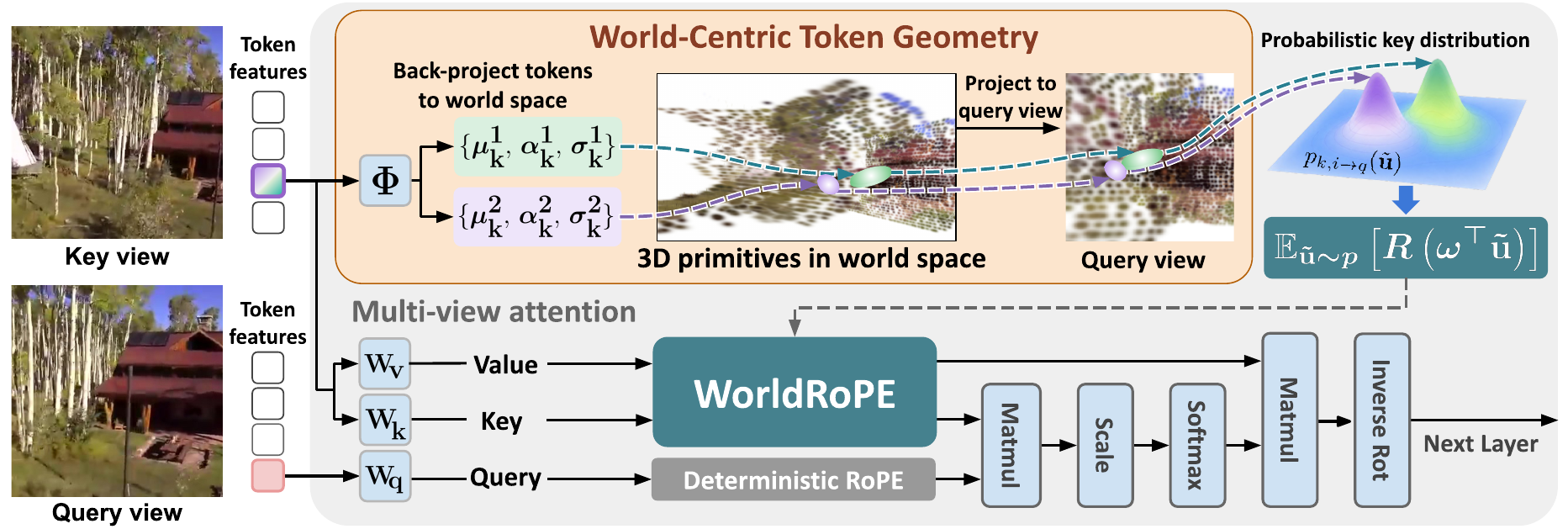}
    \caption{
    Multi-view attention with \ourmethod{}.
    At each attention layer, we predict world-space Gaussian primitives from token features.
    Projecting them into the query view places key and query positions in a common image frame.
    We integrate RoPE over the projected key distribution to guide cross-view attention, and the updated features guide geometry prediction in the next layer.
    }
    \label{fig:multi_view_att_pipeline}
    \vspace{-2mm}
\end{figure}

\paragraph{Multi-hypothesis depth prediction.}
At each layer, a prediction head $\Phi$ estimates a depth distribution along the camera ray passing through the patch center.
We model this distribution as an $M$-component Gaussian mixture with parameters given by
\begin{align}
    \left\{
    (\alpha_{k,i}^{m},\mu_{k,i}^{m},\sigma_{k,i}^{m})
    \right\}_{m=1}^{M}
    &=
    \Phi(\mathbf{f}_{k,i}),
    \label{eq:depth_head}
\end{align}
where $m$ indexes depth hypotheses, and $\mu_{k,i}^{m}$ and $\sigma_{k,i}^{m}$ denote the mean and standard deviation of the depth along the patch-center ray.
The weights satisfy $\alpha_{k,i}^{m}\geq0$ and $\sum_m\alpha_{k,i}^{m}=1$.
This mixture retains alternative depths and their uncertainty for patches containing objects at different depths.

%
%

\paragraph{Volumetric Gaussian primitives.}
An often overlooked fact is that a 2D image token corresponds to a volumetric region within the patch viewing frustum, extending beyond its center ray.
We therefore construct a Gaussian primitive around each depth hypothesis to capture both the patch extent and depth uncertainty.
Assuming independent pixel and depth offsets, we linearize back-projection to obtain the Gaussian mean and approximate covariance in the key-camera frame as follows:
\begin{align}
    \boldsymbol{\mu}_{k,i}^{m}
    &=
    \mu_{k,i}^{m}\mathbf{r}_{k,i}
    \quad \text{and} \quad
    \bm{\Sigma}_{k,i}^{m}
    \approx
    (\mu_{k,i}^{m})^2
    \mathbf{B}_k\mathbf{C}\mathbf{B}_k^\top
    +
    (\sigma_{k,i}^{m})^2
    \mathbf{r}_{k,i}\mathbf{r}_{k,i}^\top,
    \label{eq:local_cov}
\end{align}
where $\mathbf{r}_{k,i}=\mathbf{K}_k^{-1}(u_{k,i},v_{k,i},1)^\top$ and $\mathbf{B}_k$ contains the first two columns of $\mathbf{K}_k^{-1}$.
For uniform pixel offsets within a patch of size $\Delta_u\times\Delta_v$,
$\mathbf{C}=\operatorname{diag}(\Delta_u^2/12,\Delta_v^2/12)$.
The two covariance terms capture the patch extent and depth uncertainty, respectively.
See Appendix~\ref{sec:gaussian_construction} for the derivation.

%
%
%
%
%
%

\paragraph{World-space Gaussian support.}
We transform the primitives into a shared 3D world space using the key-camera extrinsics $(\mathbf{R}_k,\mathbf{t}_k)$ as follows:
\begin{align}
    \boldsymbol{\mu}_{k,i}^{\mathrm w,m}
    &=
    \mathbf{R}_k^\top
    (\boldsymbol{\mu}_{k,i}^{m}-\mathbf{t}_k)
    \quad \text{and} \quad
    \bm{\Sigma}_{k,i}^{\mathrm w,m}
    =
    \mathbf{R}_k^\top
    \bm{\Sigma}_{k,i}^{m}
    \mathbf{R}_k,
    \label{eq:world_cov}
\end{align}
where $\boldsymbol{\mu}_{k,i}^{\mathrm w,m}$ and $\bm{\Sigma}_{k,i}^{\mathrm w,m}$ are the world-space mean and covariance.
The primitives share the token feature $\mathbf{f}_{k,i}$, from which the key and value vectors are computed, and retain their mixture weights $\alpha_{k,i}^{m}$.
Together, they represent multiple plausible 3D supports for the token, with positional density given by
\begin{align}
    p_{k,i}^{\mathrm w}(\mathbf{x})
    &=
    \sum_{m=1}^{M}
    \alpha_{k,i}^{m}
    \Normal\!\left(
    \mathbf{x};
    \boldsymbol{\mu}_{k,i}^{\mathrm w,m},
    \bm{\Sigma}_{k,i}^{\mathrm w,m}
    \right),
\end{align}
where $\mathbf{x}\in\mathbb{R}^3$ denotes a world-space position.
This mixture probabilistically represents where a 2D image token may lie in 3D world space using a set of volumetric Gaussian primitives.

\subsection{Probabilistic Rotary Positional Embedding}
\label{subsection:token_splat}

%
%
%
%
%

\paragraph{Projection into the query view.}
We project the world-space Gaussian primitives into the query view to determine where each key token may appear.
Following 3D Gaussian Splatting~\citep{kerbl20233d}, a local affine approximation yields the projected mean and covariance as follows:
\begin{align}
    \boldsymbol{\mu}_{k,i\to q}^{m}
    &=
    \pi_q\!\left(
    \mathbf{R}_q\boldsymbol{\mu}_{k,i}^{\mathrm w,m}
    +\mathbf{t}_q
    \right),
    \qquad
    \bm{\Sigma}_{k,i\to q}^{m}
    =
    \mathbf{J}_{k,i\to q}^{m}
    \mathbf{R}_q
    \bm{\Sigma}_{k,i}^{\mathrm w,m}
    \mathbf{R}_q^\top
    (\mathbf{J}_{k,i\to q}^{m})^\top,
    \label{eq:projected_gaussian}
\end{align}
where $\pi_q$ denotes perspective projection from query-camera coordinates to pixel coordinates using $\mathbf{K}_q$.
Its Jacobian $\mathbf{J}_{k,i\to q}^{m}\in\mathbb{R}^{2\times3}$ is computed at the primitive center in the query-camera frame.
The projected primitives form a Gaussian mixture with positional density given by
\begin{align}
    p_{k,i\to q}(\tilde{\mathbf{u}})
    &=
    \sum_{m=1}^{M}
    \alpha_{k,i}^{m}
    \Normal\!\left(
    \tilde{\mathbf{u}};
    \boldsymbol{\mu}_{k,i\to q}^{m},
    \bm{\Sigma}_{k,i\to q}^{m}
    \right),
    \label{eq:projected_mixture}
\end{align}
where $\tilde{\mathbf{u}}\in\mathbb{R}^2$ denotes a possible projected key position in the query image.
This distribution captures the possible key locations and their uncertainty without rasterizing or warping token features.

\paragraph{Closed-form rotary integration.}
Unlike the deterministic positions used in standard RoPE, our projected key positions are probabilistic.
We therefore integrate the rotary operator over the projected distribution, defining
$\overline{\mathcal{R}}_{k,i\to q}
:=\E_{\tilde{\mathbf{u}}\sim p_{k,i\to q}}
[\mathcal{R}(\tilde{\mathbf{u}})]$.
We compute this expectation for each $2\times2$ rotation block, $R(\bm{\omega}^\top\tilde{\mathbf{u}})$; applying the result independently to all blocks yields the full operator $\mathcal{R}$.
For a fixed frequency $\bm{\omega}$, the Gaussian expectation is given by
\begin{align}
\E_{\tilde{\mathbf{u}}\sim
\Normal(\boldsymbol{\mu}_{k,i\to q}^{m},
\bm{\Sigma}_{k,i\to q}^{m})}
[R(\bm{\omega}^\top\tilde{\mathbf{u}})]
&=
\int_{\mathbb{R}^2}
R(\bm{\omega}^\top\tilde{\mathbf{u}})
\Normal\!\left(
\tilde{\mathbf{u}};
\boldsymbol{\mu}_{k,i\to q}^{m},
\bm{\Sigma}_{k,i\to q}^{m}
\right)
\,d\tilde{\mathbf{u}}
\nonumber\\
&=
\underbrace{
\exp\!\left(
-\tfrac12\bm{\omega}^\top
\bm{\Sigma}_{k,i\to q}^{m}\bm{\omega}
\right)
}_{a_{k,i\to q}^{m}}
\underbrace{
R(\bm{\omega}^\top\boldsymbol{\mu}_{k,i\to q}^{m})
}_{\text{rotation at the projected mean}}.
\label{eq:splat_integral}
\end{align}
Each primitive contributes a rotation at its projected center with covariance-dependent attenuation $a_{k,i\to q}^{m}$.
By linearity of expectation, the mixture combines these contributions into an expected block, which can be written as a single scaled rotation as follows:
\begin{align}
    \overline{R}_{k,i\to q}
    &=
    \sum_{m=1}^{M}
    \alpha_{k,i}^{m}a_{k,i\to q}^{m}
    R(\bm{\omega}^\top\boldsymbol{\mu}_{k,i\to q}^{m})
    =
    \bar a_{k,i\to q}
    R(\bar\theta_{k,i\to q}),
    \label{eq:mixture_operator}
\end{align}
where $\bar a_{k,i\to q}$ and $\bar\theta_{k,i\to q}$ denote the effective attenuation and phase for the current frequency block.
Placing these blocks along the diagonal yields the full operator $\overline{\mathcal{R}}_{k,i\to q}\in\mathbb{R}^{d\times d}$.
This expectation is exact under the projected Gaussian approximation.
See Appendix~\ref{app:sensitivity_bound} for the derivation.

\paragraph{Multi-view attention.}
We apply RoPE to the query at its deterministic coordinate $\mathbf{u}_{q,j}$ and the expected rotary operator to each key in query view $q$, as shown in Figure~\ref{fig:multi_view_att_pipeline}.
The attention score and weight are given by
\begin{align}
    s_{ji}^{q\leftarrow k}
    &=
    \frac{
    \left\langle
    \mathcal{R}(\mathbf{u}_{q,j})\mathbf{q}_{q,j},
    \overline{\mathcal{R}}_{k,i\to q}\mathbf{k}_{k,i}
    \right\rangle
    }{\sqrt{d}},
    \qquad
    A_{ji}^{q\leftarrow k}
    =
    \operatorname{softmax}_{(k,i)}
    \!\left(s_{ji}^{q\leftarrow k}\right),
    \label{eq:logit}
\end{align}
where softmax normalizes over all attended key tokens $(k,i)$ across views.
Following GTA~\citep{miyato2024gta}, we also apply positional transformations to value and output features, as detailed in Appendix~\ref{app:vo_output}.
The features updated via attention then guide geometry prediction at the next layer.

\subsection{Properties of \ourmethod{}}
\label{subsection:principle_analysis}

\paragraph{Token-adaptive temperature.}
The inverse of the effective attenuation can be interpreted as a token-adaptive temperature for each rotary frequency block.
Let $\mathbf{q},\mathbf{k}\in\mathbb{R}^2$ denote the query and key channel pairs at frequency $\bm{\omega}$.
For $\bar a_{k,i\to q}>0$, their score contribution can be rewritten:
\begin{align}
    \frac{
    \left\langle
    R(\bm{\omega}^\top\mathbf{u}_{q,j})\mathbf{q},
    \overline{R}_{k,i\to q}\mathbf{k}
    \right\rangle
    }{\sqrt{d}}
    &=
    \frac{
    \left\langle
    R(\bm{\omega}^\top\mathbf{u}_{q,j})\mathbf{q},
    R(\bar\theta_{k,i\to q})\mathbf{k}
    \right\rangle
    }{\sqrt{d}\,\tau_{k,i\to q}},
    \label{eq:adaptive_temperature_attention}
\end{align}
where $\tau_{k,i\to q}=1/\bar a_{k,i\to q}$ acts as a token-adaptive temperature.
The effective attenuation depends on the covariance of each hypothesis; 
a smaller $\bar a_{k,i\to q}$ corresponds to a higher temperature, reducing the magnitude of the score contribution for that frequency block.

\paragraph{Sensitivity to positional displacement.}
Our covariance-dependent attenuation reduces sensitivity to errors in the projected centers of primitives.
Consider an arbitrary projected Gaussian $\Normal(\boldsymbol{\mu},\bm{\Sigma})$ with attenuation $a=\exp(-\bm{\omega}^\top\bm{\Sigma}\bm{\omega}/2)$ at frequency $\bm{\omega}$.
The operator change under a center displacement $\boldsymbol{\delta}\in\mathbb{R}^2$ is bounded as follows:
\begin{align}
    \left\|
    aR\!\left(\bm{\omega}^\top(\boldsymbol{\mu}+\boldsymbol{\delta})\right)
    -aR(\bm{\omega}^\top\boldsymbol{\mu})
    \right\|_2
    &\leq
    a\left|\bm{\omega}^\top\boldsymbol{\delta}\right|,
    \label{eq:mean_sensitivity}
\end{align}
where $\|\cdot\|_2$ denotes the spectral norm.
Greater positional uncertainty, measured by $\bm{\omega}^\top\bm{\Sigma}\bm{\omega}$, decreases the attenuation factor $a$.
This lowers the upper bound on sensitivity to projected-center errors.
See Appendix~\ref{app:sensitivity_bound} for the derivation.



\paragraph{Deterministic RoPE as a limiting case.}
Our formulation recovers deterministic RoPE when one hypothesis $m^\star$ has unit weight and its patch extent and depth uncertainty vanish as follows:
\begin{align}
\alpha_{k,i}^{m^\star}\to 1,\quad
\bm{\Sigma}_{k,i\to q}^{m^\star}\to\mathbf{0}
\quad\Longrightarrow\quad
\overline{\mathcal{R}}_{k,i\to q}
\to
\mathcal{R}(\boldsymbol{\mu}_{k,i\to q}^{m^\star}).
\label{eq:deterministic_limit}
\end{align}
Thus, \ourmethod{} extends rotary encoding to probabilistic positional distributions while retaining deterministic RoPE as a limiting case.

\section{Experiments}
\label{sec:experiments}

We first evaluate \ourmethod{} for novel view synthesis against existing positional encodings.
We then conduct detailed analyses of \ourmethod{}, including ablation studies.
Finally, we demonstrate that \ourmethod{} can serve as a plug-and-play mechanism for other tasks using multi-view attention.

\subsection{Main Results}

\begin{table}[t]
    \centering
    \caption{
        Comparison of positional encodings for novel view synthesis with LVSM
        on RE10K and Objaverse.
        \ourmethod{} achieves the best performance across all metrics on both datasets.
    }
    \vspace{-3mm}
    \setlength{\tabcolsep}{13pt}
    \resizebox{\linewidth}{!}{%
        \begin{tabular}{l|ccc|ccc}
            \toprule
            \multirow{2}{*}{\textbf{Method}}
            & \multicolumn{3}{c|}{\textbf{RE10K}~\citep{zhou2018stereo}}
            & \multicolumn{3}{c}{\textbf{Objaverse}~\citep{deitke2023objaverse}} \\
            & \textbf{PSNR} $\uparrow$
            & \textbf{SSIM} $\uparrow$
            & \textbf{LPIPS} $\downarrow$
            & \textbf{PSNR} $\uparrow$
            & \textbf{SSIM} $\uparrow$
            & \textbf{LPIPS} $\downarrow$ \\
            \midrule
            Pl\"ucker Ray~\citep{jin2024lvsm}
            & 24.62 & 0.790 & 0.108
            & 15.15 & 0.838 & 0.355 \\

            GTA~\citep{miyato2024gta}
            & 25.05 & 0.799 & 0.102
            & 22.22 & 0.892 & 0.123 \\

            PRoPE~\citep{li2026prope}
            & 25.44 & 0.811 & 0.095
            & 22.41 & 0.894 & 0.117 \\

            RayRoPE~\citep{wu2026rayrope}
            & 26.17 & 0.833 & 0.081
            & 22.25 & 0.892 & 0.121 \\

            URoPE~\citep{xie2026urope}
            & 26.26 & 0.834 & 0.078
            & 21.99 & 0.890 & 0.128 \\
            \midrule

            \ourmethod{} (ours)
            & \textbf{26.75}
            & \textbf{0.846}
            & \textbf{0.073}
            & \textbf{22.52}
            & \textbf{0.896}
            & \textbf{0.113} \\
            \bottomrule
        \end{tabular}%
    }
    \label{tab:lvsm_quantitative_results}
\end{table}

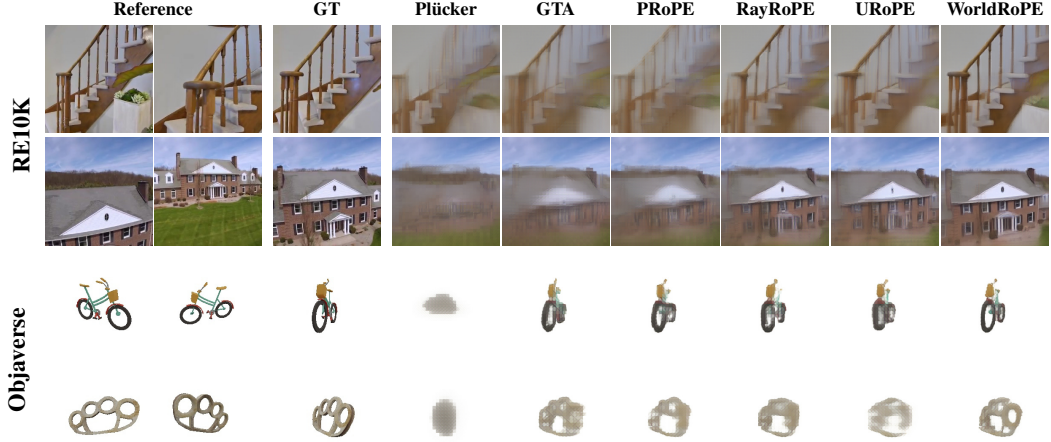
\begin{figure*}[t]
\centering
\vspace{-1mm}
\begin{tikzpicture}
    \def\qualTopGap{-2pt}   
    \def\qualSideGap{2pt}  

    \node[anchor=south west, inner sep=0] (img)
        at (0.05\textwidth, 0)
        {\includegraphics[width=0.95\textwidth]{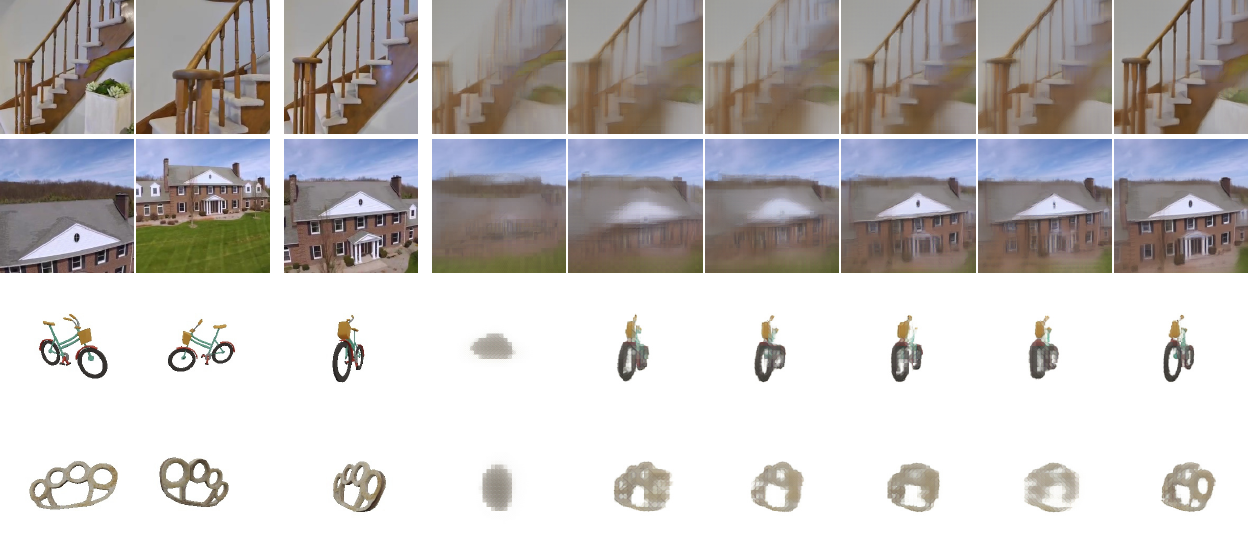}};

    \foreach \x/\name in {
        0.10810/{Reference},
        0.28130/{GT},
        0.39983/{Pl\"ucker},
        0.50918/{GTA},
        0.61853/{PRoPE},
        0.72788/{RayRoPE},
        0.83723/{URoPE},
        0.94658/{\ourmethod{}}
    } {
        \node[anchor=south, font=\scriptsize\bfseries]
            at ($(img.north west)!\x!(img.north east)+(0,\qualTopGap)$)
            {\strut\name};
    }

    \node[rotate=90, anchor=south, font=\footnotesize\bfseries]
        at ($(img.south west)!0.75188!(img.north west)
            +(-\qualSideGap,0)$) {RE10K};
    \node[rotate=90, anchor=south, font=\footnotesize\bfseries]
        at ($(img.south west)!0.24812!(img.north west)
            +(-\qualSideGap,0)$) {Objaverse};
\end{tikzpicture}
    \vspace{-7.5mm}
    \caption{
    Qualitative comparison of novel view synthesis on RE10K and Objaverse. \ourmethod{} better preserves thin structures and fine details like stair railings and bicycle wheels. 
    }
    \label{fig:lvsm_qualitative_results}
    \vspace{-2mm}
\end{figure*}

\paragraph{Experimental setup.}
We test \ourmethod{} within LVSM~\citep{jin2024lvsm} on RE10K~\citep{zhou2018stereo} and Objaverse~\citep{deitke2023objaverse}.
We compare with Pl\"ucker ray encoding~\citep{jin2024lvsm}, GTA~\citep{miyato2024gta}, PRoPE~\citep{li2026prope}, RayRoPE~\citep{wu2026rayrope}, and URoPE~\citep{xie2026urope}, using PSNR, SSIM, and LPIPS~\citep{zhang2018perceptual} to measure rendered image quality.


\vspace{-2mm}
\paragraph{Quantitative results.}
Table~\ref{tab:lvsm_quantitative_results} shows that \ourmethod{} consistently outperforms all baselines across all metrics on both the RE10K dataset of real-world scenes and the Objaverse dataset of object-centric scenes.
The improvement is particularly pronounced on the RE10K dataset, with a PSNR gain of $0.49$ dB over the strongest baseline.
This suggests that modeling alternative depths and their spatial extent is particularly useful for scenes containing foreground and background structures, where a patch can span multiple depths and correspond to a broader region in another view.

\vspace{-2mm}
\paragraph{Qualitative results.}
Figure~\ref{fig:lvsm_qualitative_results} compares synthesized target views.
\ourmethod{} better preserves thin structures and fine details, including stair railings, window boundaries, bicycle wheels, and brass knuckles, which the baselines often blur or merge with their surroundings.
These observations suggest that modeling alternative depth hypotheses helps distinguish structures at different depths within a patch.
The qualitative improvements are consistent with the gains in PSNR, SSIM, and LPIPS.

\subsection{Analysis and Discussion}
\label{sec:experimental_analysis}

\paragraph{Component ablation.}
Table~\ref{tab:component_ablation} separates the effects of probabilistic modeling, multiple depth hypotheses, and the patch footprint.
Removing probabilistic modeling causes the largest drop, indicating the importance of incorporating positional uncertainty into attention.
Performance also declines with a single hypothesis, supporting the representation of alternative depths within a patch.
The smaller but consistent drop without the patch footprint shows that its spatial extent contributes beyond the patch-center ray.


\vspace{-2mm}
\paragraph{Number of hypotheses.}
Table~\ref{tab:mixture_ablation} tests how many depth hypotheses are needed per token.
Increasing the number from one to four improves all three metrics, supporting the use of alternative depths within a patch.
Eight hypotheses bring no further gain; we therefore use four primitives per token.

\begin{table*}[t]
\centering
\begin{minipage}[t]{0.485\textwidth}
\vspace{0pt}
\centering
\caption{
Component ablation on RE10K.
Removing any component degrades performance; modeling with probabilistic primitives yields the largest performance gain.
}
\label{tab:component_ablation}
\vspace{-2mm}
\scalebox{0.78}{%
    \setlength{\tabcolsep}{5pt}
    \begin{tabular}{l|ccc}
        \toprule
        \textbf{Method}
        & \textbf{PSNR} $\uparrow$
        & \textbf{SSIM} $\uparrow$
        & \textbf{LPIPS} $\downarrow$ \\
        \midrule
        w/o Probabilistic modeling
        & 26.30 & 0.835 & 0.082 \\
        w/o Multi-hypothesis
        & 26.50 & 0.839 & 0.077 \\
        w/o Patch footprint
        & 26.68 & 0.845 & 0.074 \\
        \midrule
        \ourmethod{}
        & \textbf{26.75} & \textbf{0.846} & \textbf{0.073} \\
        \bottomrule
    \end{tabular}%
}
\end{minipage}
\hfill
\begin{minipage}[t]{0.485\textwidth}
\vspace{0pt}
\centering
\caption{
Effect of the number of hypotheses on RE10K.
Performance improves from one to four hypotheses, while eight brings no further gain.
We use four hypotheses per token.
}\label{tab:mixture_ablation}
\vspace{-2mm}
\scalebox{0.78}{%
    \setlength{\tabcolsep}{8.3pt}
    \begin{tabular}{c|ccc}
        \toprule
        \textbf{\# of hypotheses}
        & \textbf{PSNR} $\uparrow$
        & \textbf{SSIM} $\uparrow$
        & \textbf{LPIPS} $\downarrow$ \\
        \midrule
        1 & 26.50 & 0.839 & 0.077 \\
        2 & 26.61 & 0.842 & 0.075 \\
        4 & \textbf{26.75} & \textbf{0.846} & \textbf{0.073} \\
        8 & 26.73 & \textbf{0.846} & \textbf{0.073} \\
        \bottomrule
    \end{tabular}%
}
\end{minipage}
\vspace{1mm}
\end{table*}

\vspace{-2mm}
\paragraph{Visualization of depth hypotheses.}
\begin{figure}[t]
    \centering
    \begin{subfigure}[t]{0.4771\linewidth}
        \vspace{0pt}
        \centering
        \captionsetup{skip=2pt}
        \includegraphics[width=\linewidth]
            {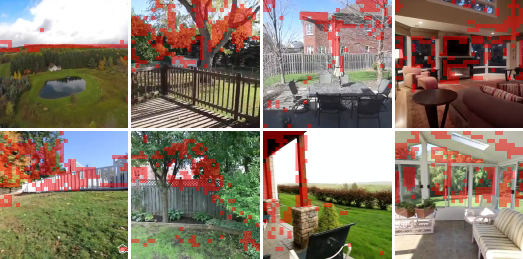}
        \caption{Tokens with multiple depth peaks}
        \vspace{-1mm}
        \label{fig:multi-depth_a}
    \end{subfigure}%
    \hspace{0.04\linewidth}%
    \begin{subfigure}[t]{0.4829\linewidth}
        \vspace{0pt}
        \centering
        \captionsetup{skip=2pt}
        \includegraphics[
            width=\linewidth,
            trim={0 3.35pt 0 3.68pt},
            clip
        ]{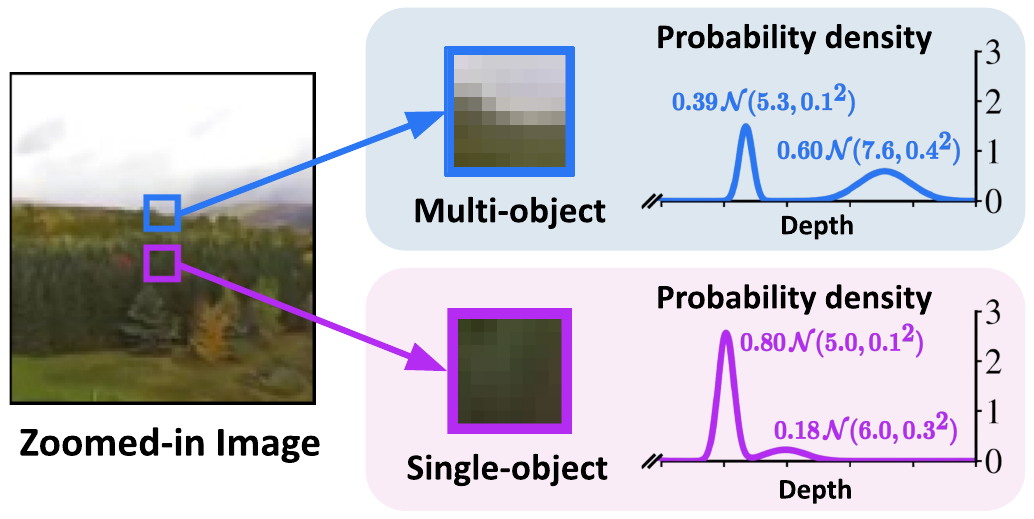}
        \caption{Predicted depth distributions for each token}
        \vspace{-1mm}
        \label{fig:multi-depth_b}
    \end{subfigure}
    \caption{
        Visualization of predicted depth hypotheses.
        (a) Red highlights tokens with multiple prominent peaks in their
        depth distributions.
        (b) A patch spanning vegetation and sky exhibits two distinct peaks,
        while a patch within vegetation has one dominant peak.
    }
    \vspace{-4mm}
    \label{fig:multi-depth}
\end{figure}

Figure~\ref{fig:multi-depth} examines how predicted depth hypotheses vary with patch content.
In Figure~\ref{fig:multi-depth_a}, red highlights tokens with multiple peaks, which often appear near object boundaries and overlapping structures.
A patch spanning vegetation and sky produces two peaks, whereas a patch within vegetation produces one dominant peak, as shown in Figure~\ref{fig:multi-depth_b}.
This contrast shows that \ourmethod{} can preserve distinct depths within a patch instead of collapsing them into a single estimate.

\vspace{-2mm}
\paragraph{Compatibility with other rotary operators.}

\begin{figure}[t]
\centering
\begin{minipage}[t]{0.52\linewidth}
\vspace{0pt}
\centering
\begin{subfigure}[t]{0.49\linewidth}
    \centering
    \includegraphics[width=0.495586\linewidth]
        {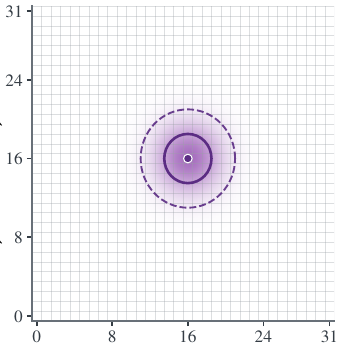}%
    \hfill
    \includegraphics[width=0.464414\linewidth]
        {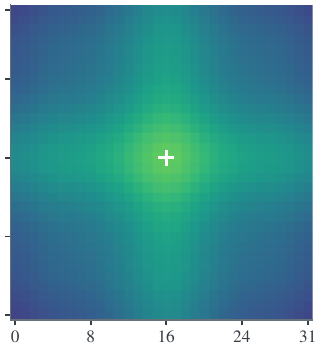}%
    \par\vspace{1mm}
    \includegraphics[width=0.495586\linewidth]
        {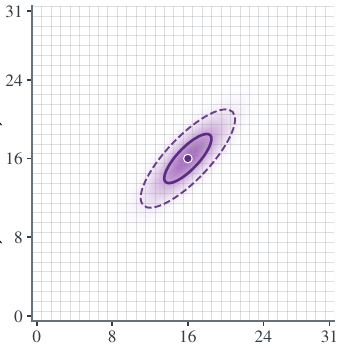}%
    \hfill
    \includegraphics[width=0.464414\linewidth]
        {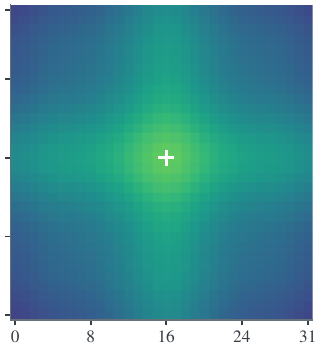}
    \caption{Axial rotary}
    \label{fig:axial_response}
\end{subfigure}%
\hfill
\begin{subfigure}[t]{0.49\linewidth}
    \centering
    \includegraphics[width=0.495586\linewidth]
        {asset/rotary_operators/sigma0_projected_gaussian.pdf}%
    \hfill
    \includegraphics[width=0.464414\linewidth]
        {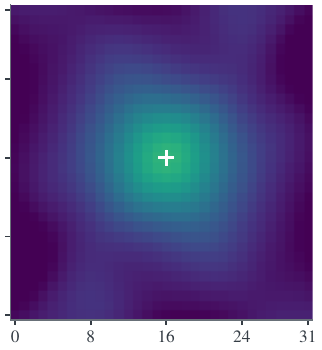}%
    \par\vspace{1mm}
    \includegraphics[width=0.495586\linewidth]
        {asset/rotary_operators/sigma5_projected_gaussian.pdf}%
    \hfill
    \includegraphics[width=0.464414\linewidth]
        {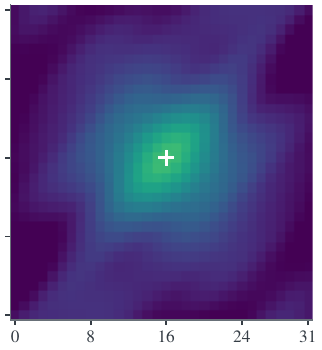}
    \caption{Mixed rotary}
    \label{fig:mixed_response}
\end{subfigure}
\vspace{-2mm}
\caption{
Projected Gaussians and their responses under axial and mixed rotary operators.
Only the mixed operator responds to off-diagonal covariance.
}
\label{fig:mixed_rope_comparison}
\end{minipage}%
\hfill
\begin{minipage}[t]{0.45\linewidth}
\vspace{0pt}
\centering
\captionof{table}{
PSNR with axial and mixed rotary operators on Objaverse.
Parenthesized values denote changes from the axial variant.
The mixed operator yields the largest gain for \ourmethod{}.
Additional results are reported in Appendix~\ref{subsec:mixedrope}.
}
\label{tab:main_mixed_rope}
\vspace{-2mm}
\renewcommand{\arraystretch}{1.08}
\scalebox{0.78}{%
    \setlength{\tabcolsep}{4pt}
    \begin{tabular}{@{}lll@{}}
    \toprule
        \textbf{Method}
        & \textbf{Operator}
    &~~~~~ \textbf{PSNR} $\uparrow$ \\
    \midrule
    \multirow{2}{*}{RayRoPE~\citep{wu2026rayrope}}
    & Axial & 22.25~~~~~~~~~~~~~~~ \\
    & Mixed & 22.33 \textcolor{blue}{$\bm{(+0.08)}$} \\
    \midrule
    \multirow{2}{*}{URoPE~\citep{xie2026urope}}
    & Axial & 21.99 ~~~~~~~~~~~~~~~\\
    & Mixed & 22.03 \textcolor{blue}{$\bm{(+0.04)}$} \\
    \midrule
    \multirow{2}{*}{\ourmethod{} (ours)}
    & Axial & \textbf{22.52} ~~~~~~~~~~~~~~~ \\
    & Mixed & \textbf{22.71} \textcolor{blue}{$\bm{(+0.19)}$} \\
    \bottomrule
    \end{tabular}%
}
\end{minipage}
\end{figure}

To examine how the rotary operator uses our predicted Gaussian covariance, we replace the axial operator used in the main experiments with a mixed operator.
Axial operators use frequency vectors aligned with one image axis, whereas mixed operators combine both coordinates.\footnote{Axial operators use $\bm{\omega}=(\omega_u,0)^\top$ or $(0,\omega_v)^\top$, whereas mixed operators use $\bm{\omega}=(\omega_u,\omega_v)^\top$.}
In the Gaussian expectation derived in Section~\ref{subsection:token_splat}, this distinction allows only mixed operators to use covariance between the image coordinates, as illustrated in Figures~\ref{fig:axial_response} and~\ref{fig:mixed_response}.
On the Objaverse dataset, the mixed operator improves \ourmethod{} by $0.19$ dB in PSNR, compared with $0.08$ dB for RayRoPE and $0.04$ dB for URoPE (Table~\ref{tab:main_mixed_rope}).
The larger gain suggests that incorporating off-diagonal covariance into the rotary response makes our Gaussian modeling more effective.
See Appendix~\ref{subsec:mixedrope} for results on the RE10K dataset.

\begin{figure}[t]
    \centering
    \vspace{-1mm}
    \includegraphics[width=\linewidth]
        {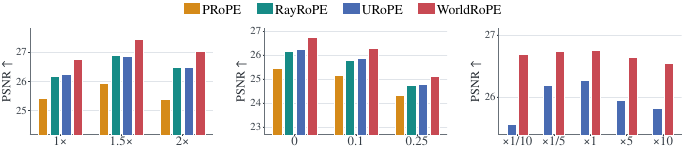}
    \par\vspace{-5mm}
    \makebox[\linewidth][l]{%
        \hspace*{0.0465\linewidth}%
        \subcaptionbox{
         Zoom-in
                     \label{fig:re10k_zoom_psnr}
        }[0.285\linewidth][c]{\rule{0pt}{0pt}}%
        \hspace*{0.0520\linewidth}%
        \subcaptionbox{
             Rotation noise
                 \label{fig:re10k_rotation_psnr}
        }[0.285\linewidth][c]{\rule{0pt}{0pt}}%
        \hspace*{0.0520\linewidth}%
        \subcaptionbox{
            Depth initialization
                \label{fig:depth_initialization}
        }[0.285\linewidth][c]{\rule{0pt}{0pt}}%
    }
    \vspace{-2mm}
    \caption{
    Robustness to (a) camera zoom and (b) rotation noise during evaluation, and (c) depth configuration during training.
    \ourmethod{} achieves robustness across all tested settings.
    }
    \label{fig:re10k_ood_psnr}
    \vspace{-5mm}
\end{figure}

\vspace{-2mm}
\paragraph{Robustness.}
We test sensitivity to camera perturbations during evaluation and depth initialization during training.
Under camera zoom and rotation noise, \ourmethod{} achieves higher PSNR than the baselines (Figures~\ref{fig:re10k_zoom_psnr} and~\ref{fig:re10k_rotation_psnr}).
Zoom changes the projected extent of a patch, whereas rotation noise shifts its projected position.
These results are consistent with the benefits of modeling the patch footprint and accounting for positional uncertainty.
We also scale the predefined depth anchors of URoPE and the initial depth biases of \ourmethod{} from $0.1$ to $10$ times their default values.
\ourmethod{} maintains higher PSNR with less variation across scales (Figure~\ref{fig:depth_initialization}).
Its depth hypotheses are updated from token features during training, reducing dependence on their initial placement.

\begin{figure}[t]
    \centering
    \vspace{-2mm}
    \includegraphics[width=0.99\linewidth]{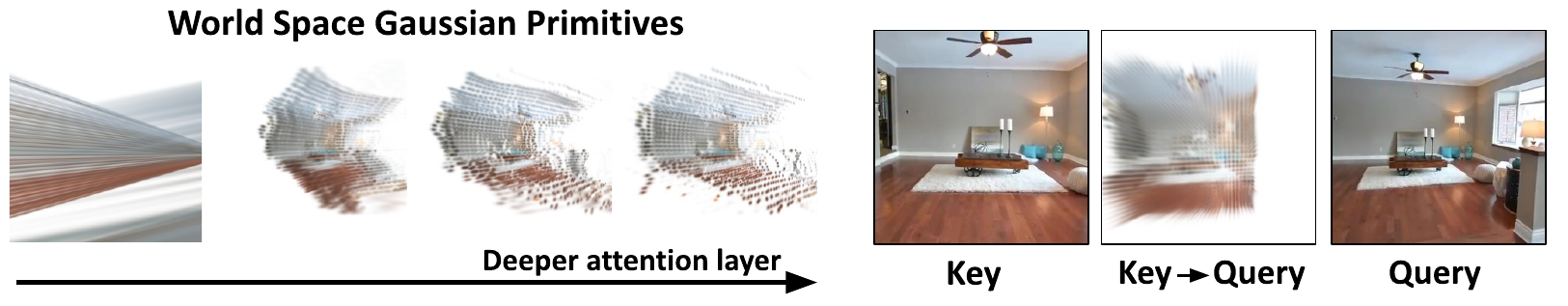}
    \vspace{-2mm}
     \caption{
    World-space Gaussian primitives across attention layers (left) and their projection from the key view into the query view (right).
    RGB rasterization is used only for visualization; the primitives carry token features and are not rasterized during attention.}
    \label{fig:GS_layerwise_mainpaper}
\end{figure}

\vspace{-2mm}
\paragraph{Layerwise geometry refinement.}
Figure~\ref{fig:GS_layerwise_mainpaper} shows the predicted Gaussian primitives across attention layers and their projection into the query view.
In deeper layers, their smaller extents indicate reduced positional uncertainty, while the scene structure becomes more recognizable.
The projected key primitives approximately align with the query image, although their centers no longer follow a regular image grid and their regions vary in size and orientation.
Some regions are broader near image boundaries, consistent with perspective magnification of a key-view patch when projected into the query view.
See Appendix~\ref{sec:supp_additional_layer_analysis} for additional visualizations across all attention layers.    

\subsection{Broader Applications}
We further show that \ourmethod{} can serve as a plug-and-play mechanism for camera-conditioned multi-view attention, including depth estimation and feed-forward 3D reconstruction.

\vspace{-2mm}
\paragraph{Depth estimation.}
We integrate camera-aware positional encodings into the cross-attention layers of UniMatch~\citep{xu2023unifying} and compare their depth estimation performance.
We train them on the training splits of the Scenes11~\citep{ummenhofer2017demon}, SUN3D~\citep{xiao2013sun3d}, and RGBD~\citep{sturm2012benchmark} datasets, and evaluate on their test splits, following UniMatch.
We further evaluate cross-dataset generalization on the unseen ScanNet++ dataset~\citep{yeshwanth2023scannet++}.
Table~\ref{tab:unimatch_depth_results} shows that \ourmethod{} achieves the lowest absolute relative error (Abs Rel) and root mean squared error (RMSE) for depth estimation across all four datasets.
The improvements in both metrics on the unseen dataset further support generalization to unseen scenes. 
Further implementation details and qualitative results are provided in Sections~\ref{sec:experimental_details_unimatch} and~\ref{sec:unimatch_qual}, respectively.

\begin{table}[t]
\centering
\caption{
Comparison of camera-aware positional encodings for stereo depth estimation with UniMatch.
We report depth errors using AbsRel and RMSE on the Scenes11, SUN3D, and RGBD datasets, and assess cross-dataset generalization on the unseen ScanNet++ dataset.
\ourmethod{} outperforms existing positional encodings in RMSE on all four datasets.
}
\vspace{-2mm}
\scalebox{0.78}{%
    \setlength{\tabcolsep}{4pt}
    \begin{tabular}{l|cc|cc|cc|cc}
        \toprule
        \multirow{2}{*}{\textbf{Method}}
        & \multicolumn{2}{c|}{\textbf{Scenes11}}
        & \multicolumn{2}{c|}{\textbf{SUN3D}}
        & \multicolumn{2}{c|}{\textbf{RGBD}}
        & \multicolumn{2}{c}{\textbf{ScanNet++ (unseen)}} \\
        & \textbf{Abs Rel} $\downarrow$
        & \textbf{RMSE} $\downarrow$
        & \textbf{Abs Rel} $\downarrow$
        & \textbf{RMSE} $\downarrow$
        & \textbf{Abs Rel} $\downarrow$
        & \textbf{RMSE} $\downarrow$
        & \textbf{Abs Rel} $\downarrow$
        & \textbf{RMSE} $\downarrow$ \\
        \midrule
        w/o camera encoding
        & 0.176 & 1.182
        & 0.179 & 0.524
        & 0.185 & 0.953
        & 0.286 & 0.482 \\
        \midrule
        Pl\"ucker Ray~\citep{jin2024lvsm}
        & 0.123 & 0.943
        & 0.130 & 0.391
        & 0.145 & 0.790
        & 0.276 & 0.464 \\
        GTA~\citep{miyato2024gta}
        & 0.116 & 0.830
        & 0.133 & 0.384
        & 0.134 & 0.693
        & 0.325 & 0.520 \\
        PRoPE~\citep{li2026prope}
        & 0.106 & 0.778
        & 0.136 & 0.390
        & 0.134 & 0.695
        & 0.263 & 0.434 \\
        RayRoPE~\citep{wu2026rayrope}
        & 0.101 & 0.743
        & 0.125 & 0.370
        & 0.128 & 0.677
        & 0.257 & 0.426 \\
        URoPE~\citep{xie2026urope}
        & 0.093 & 0.686
        & 0.123 & 0.374
        & \textbf{0.119} & 0.637
        & 0.234 & 0.398 \\
        \midrule
        \ourmethod{} (ours)
        & \textbf{0.079} & \textbf{0.661}
        & \textbf{0.116} & \textbf{0.356}
        & \textbf{0.119} & \textbf{0.630}
        & \textbf{0.224} & \textbf{0.386} \\
        \bottomrule
    \end{tabular}%
}
\vspace{-3mm}
\label{tab:unimatch_depth_results}
\end{table}

\paragraph{Feed-forward 3D reconstruction.}%
\begin{wraptable}{r}{0.45\linewidth}
    \vspace{-4mm}
    \centering
    \caption{
    Comparison of camera-aware positional encodings for feed-forward 3D reconstruction with VGGT. WorldRoPE achieves the lowest depth, camera-space point, and camera rotation errors.
    }
    \label{tab:vggt_results}
    \vspace{-2mm}
    \scalebox{0.78}{%
        \setlength{\tabcolsep}{3.5pt}
        \begin{tabular}{l|cccc}
            \toprule
            \multirow{2}{*}{\textbf{Method}}
            & \multicolumn{4}{c}{\textbf{ScanNet++}} \\
            & \shortstack{\strut\textbf{Depth}\\\strut\textbf{AbsRel} $\downarrow$}
            & \shortstack{\strut\textbf{Norm.}\\\strut\textbf{RMSE} $\downarrow$}
            & \shortstack{\strut\textbf{Cam.-pts}\\\strut\textbf{AbsRel} $\downarrow$}
            & \shortstack{\strut\textbf{Rot.}\\\strut\textbf{Err.} $\downarrow$} \\
            \midrule
            RayRoPE
            & 0.0579 & 0.0624 & 0.0575 & 1.04 \\
            URoPE
            & 0.0581 & 0.0626 & 0.0576 & 1.02 \\
            \midrule
            \ourmethod{}
            & \textbf{0.0567}
            & \textbf{0.0619}
            & \textbf{0.0565}
            & \textbf{0.98} \\
            \bottomrule
        \end{tabular}%
    }
    \vspace{-1mm}
\end{wraptable}

We fine-tune pretrained VGGT~\citep{wang2025vggt} by adding attention layers with camera-aware positional encodings.
We train on the ScanNet++~\citep{yeshwanth2023scannet++} training split and evaluate on its test split.
Table~\ref{tab:vggt_results} reports depth errors (AbsRel and normalized RMSE), camera-space point AbsRel, and camera rotation error.
\ourmethod{} outperforms baselines on all four metrics, demonstrating that our world-centric positional encoding can be effectively applied when fine-tuning a pretrained model. 
Further implementation details are shown in Appendix~\ref{sec:experimental_details_vggt}.
\vspace{-2mm}
\section{Conclusion}
\label{sec:conclusion}

We presented \ourmethod{}, a probabilistic world-centric RoPE for multi-view attention.
Our method represents token geometry with volumetric Gaussian primitives that jointly capture alternative depth hypotheses, depth uncertainty, and the spatial extent of the patch frustum.
Integrating the rotary operator over their projected distributions yields a closed-form encoding that attenuates uncertain positional interactions without rasterizing or warping token features.
Across layers, predicted geometry guides attention, while updated features inform subsequent geometry predictions.
Experiments on novel view synthesis, depth estimation, and feed-forward 3D reconstruction demonstrate the effectiveness of \ourmethod{} across multi-view tasks and architectures.

\subsection*{AI use statement}
We used generative AI tools as part of our workflow.
These tools helped us discuss methodological choices and mathematical claims, interpret experimental results, identify relevant literature, and draft and refine portions of the manuscript.
They were also used to assist with graph preparation and figure layout.
All reported results and graphs are based on experiments designed, conducted, and verified by the authors.
The authors reviewed and revised all AI-assisted material and take full responsibility for the final text, technical claims, references, figures, and experimental results.


\subsection*{Ethics statement}
This work evaluates the proposed method on existing datasets and does not introduce new data collection.
Like other multi-view reconstruction methods, it can recover scene structure and visual details from images observed across multiple viewpoints.
When applied to private or sensitive imagery, its use should respect the consent of relevant individuals, dataset licenses and usage restrictions, and appropriate privacy safeguards.
We do not anticipate additional ethical concerns beyond those associated with the source datasets and potential applications of multi-view reconstruction.


\subsection*{Reproducibility statement}
The main paper describes the proposed world-space token representation and its closed-form rotary integration.
The appendix provides further derivations, including the construction and projection of the Gaussian primitives, as well as analyses of the resulting rotary operators.
For each task, the experimental sections and appendix specify the model adaptation, training configuration, dataset splits, evaluation protocol, and relevant implementation details.
They also report ablations that isolate the effects of multi-hypothesis modeling, patch footprint, uncertainty-dependent attenuation, and the number of primitives.


\bibliography{iclr2027_conference}
\bibliographystyle{iclr2027_conference}

\appendix
\clearpage

\section{Additional Analysis}
\label{sec:additional_analysis}

\subsection{Compatibility with Mixed Rotary Operator}
\label{subsec:mixedrope}

\begin{figure}[h]
    \centering
    \begin{tikzpicture}
        \def\rowLabelGap{2pt}
        \def\mainTitleGap{1pt}

        \node[anchor=south west, inner sep=0] (img)
            at (0.07\linewidth,0)
            {\includegraphics[width=0.93\linewidth]
            {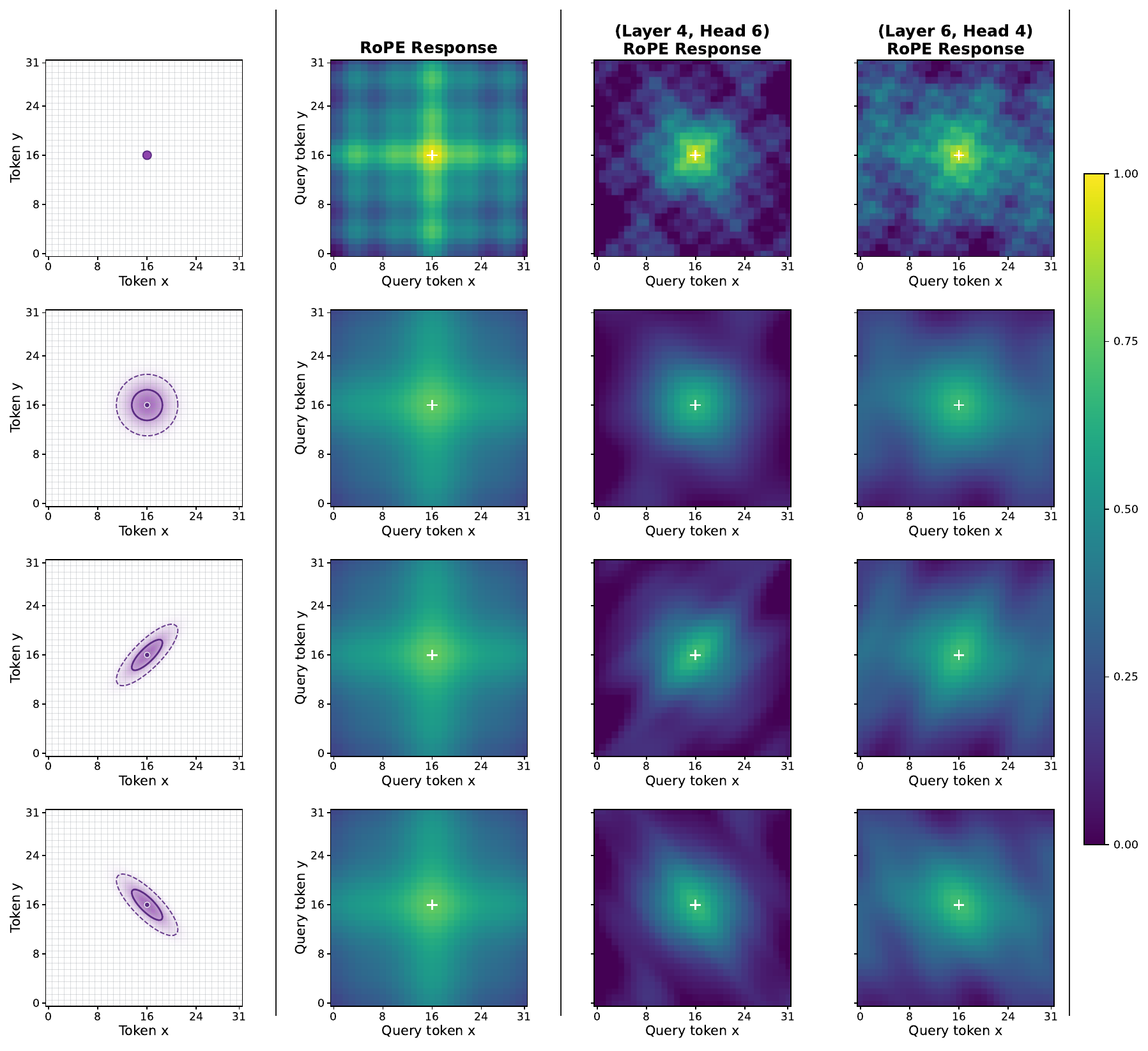}};

        \node[
            anchor=south,
            align=center,
            font=\fontsize{6}{6.5}\selectfont\bfseries
        ]
            at ($(img.north west)!0.12546!(img.north east)
                +(0,\mainTitleGap)$)
            {Projected Gaussian};

        \node[
            anchor=south,
            align=center,
            font=\fontsize{6}{6.5}\selectfont\bfseries
        ]
            at ($(img.north west)!0.37371!(img.north east)
                +(0,\mainTitleGap)$)
            {Axial RoPE};

        \node[
            anchor=south,
            align=center,
            font=\fontsize{6}{6.5}\selectfont\bfseries
        ]
            at ($(img.north west)!0.71790!(img.north east)
                +(0,\mainTitleGap)$)
            {Mixed RoPE};

        \foreach \y/\name in {
            0.84932/{(a)},
            0.61143/{(b)},
            0.37354/{(c)},
            0.13565/{(d)}
        } {
            \node[
                anchor=east,
                font=\fontsize{6}{6.5}\selectfont\bfseries
            ]
                at ($(img.south west)!\y!(img.north west)
                    +(-\rowLabelGap,0)$)
                {\name};
        }
    \end{tikzpicture}
    \caption{
Projected Gaussian supports and their responses under axial and mixed rotary operators.
We set $\mathbf{q}=\mathbf{k}=\mathbf{1}/\sqrt{d}$ to isolate the positional response.
Panel (a) shows a point hypothesis; panels (b) to (d) have the same marginal variances $\Sigma_{uu}$ and $\Sigma_{vv}$ but different off-diagonal covariance $\Sigma_{uv}$ and $\Sigma_{vu}$.
Across these panels, the axial operator gives the same response, whereas the mixed operator responds to the covariance between image coordinates.
}

    \label{fig:mixed_rope_covariance_response}
\end{figure}

Both axial and mixed rotary operators~\citep{heo2024ropevit} compute each rotation angle as
$\bm{\omega}^\top\mathbf{u}$ and thus support the Gaussian integration derived in the main paper.
They differ in the choice of $\bm{\omega}$ and the covariance terms used in attenuation.
For $\bm{\omega}=(\omega_u,\omega_v)^\top$, the phase variance is given by
\begin{align}
\bm{\omega}^\top\bm{\Sigma}\bm{\omega}
=
\omega_u^2\Sigma_{uu}
+2\omega_u\omega_v\Sigma_{uv}
+\omega_v^2\Sigma_{vv}.
\end{align}
Axial rotary operators compute each rotation angle from either the horizontal or vertical coordinate, using
$\bm{\omega}=(\omega_u,0)^\top$ or $(0,\omega_v)^\top$.
The phase variance is then given by
\begin{align}
\bm{\omega}^\top\bm{\Sigma}\bm{\omega}
=
\omega_u^2\Sigma_{uu}
\quad\text{or}\quad
\omega_v^2\Sigma_{vv},
\end{align}
so attenuation depends only on the corresponding marginal variance.
Mixed rotary operators instead combine both coordinates, allowing their attenuation to use the cross-covariance term.
They can therefore distinguish projected Gaussians with identical marginal variances but different orientations, as illustrated in Figure~\ref{fig:mixed_rope_covariance_response}.

\begin{table}[h]
\centering
\caption{
Comparison of axial and mixed rotary operators for novel view synthesis on the RE10K and Objaverse datasets.
Parenthesized values denote changes from the axial variant.
}
\renewcommand{\arraystretch}{1.08}
\setlength{\tabcolsep}{5pt}

\resizebox{\linewidth}{!}{%
\begin{tabular}{@{}llcccccc@{}}
\toprule
\multirow{2}{*}{\textbf{Method}}
& \multirow{2}{*}{\textbf{Operator}}
& \multicolumn{3}{c}{\textbf{RE10K}~\citep{zhou2018stereo}}
& \multicolumn{3}{c}{\textbf{Objaverse}~\citep{deitke2023objaverse}} \\
\cmidrule(lr){3-5}
\cmidrule(lr){6-8}
&
& \textbf{PSNR} $\uparrow$
& \textbf{SSIM} $\uparrow$
& \textbf{LPIPS} $\downarrow$
& \textbf{PSNR} $\uparrow$
& \textbf{SSIM} $\uparrow$
& \textbf{LPIPS} $\downarrow$ \\
\midrule

\multirow{2}{*}{RayRoPE~\citep{wu2026rayrope}}
& Axial
& 26.17 & 0.833 & 0.081
& 22.25 & 0.892 & 0.121 \\
& Mixed
& 26.14 {\scriptsize($-0.03$)}
& 0.832 {\scriptsize($-0.001$)}
& 0.082 {\scriptsize($+0.001$)}
& 22.33 {\scriptsize($+0.08$)}
& 0.894 {\scriptsize($+0.002$)}
& 0.118 {\scriptsize($-0.003$)} \\
\midrule

\multirow{2}{*}{URoPE~\citep{xie2026urope}}
& Axial
& 26.26 & 0.834 & 0.078
& 21.99 & 0.890 & 0.128 \\
& Mixed
& 26.27 {\scriptsize($+0.01$)}
& 0.834 {\scriptsize($0.000$)}
& 0.079 {\scriptsize($+0.001$)}
& 22.03 {\scriptsize($+0.04$)}
& 0.890 {\scriptsize($0.000$)}
& 0.126 {\scriptsize($-0.002$)} \\
\midrule

\multirow{2}{*}{\ourmethod{} (ours)}
& Axial
& 26.75 & 0.846 & 0.073
& 22.52 & 0.896 & 0.113 \\
& Mixed
& \textbf{26.84} {\scriptsize\textbf{($+0.09$)}}
& \textbf{0.848} {\scriptsize\textbf{($+0.002$)}}
& \textbf{0.072} {\scriptsize\textbf{($-0.001$)}}
& \textbf{22.71} {\scriptsize\textbf{($+0.19$)}}
& \textbf{0.898} {\scriptsize\textbf{($+0.002$)}}
& \textbf{0.108} {\scriptsize\textbf{($-0.005$)}} \\
\bottomrule
\end{tabular}%
}
\label{tab:mixed_rope_results}
\end{table}
As shown in Table~\ref{tab:mixed_rope_results}, \ourmethod{} gains most consistently from mixed rotary operators, while RayRoPE and URoPE show limited or inconsistent improvements.
This suggests that mixed rotary operators make fuller use of the cross-coordinate covariance in our projected Gaussians.

\subsection{Sensitivity to depth initialization.}

\begin{figure}[h]
    \centering
    \begin{subfigure}[b]{0.32\linewidth}
        \centering
        \includegraphics[width=\linewidth]
            {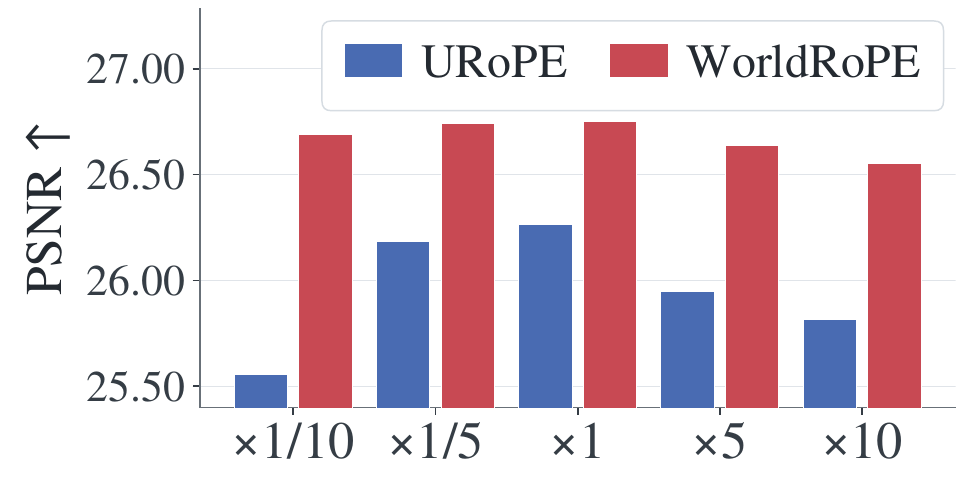}
        \caption{PSNR $\uparrow$}
        \label{fig:anchor_init_psnr}
    \end{subfigure}
    \hfill
    \begin{subfigure}[b]{0.32\linewidth}
        \centering
        \includegraphics[width=\linewidth]
            {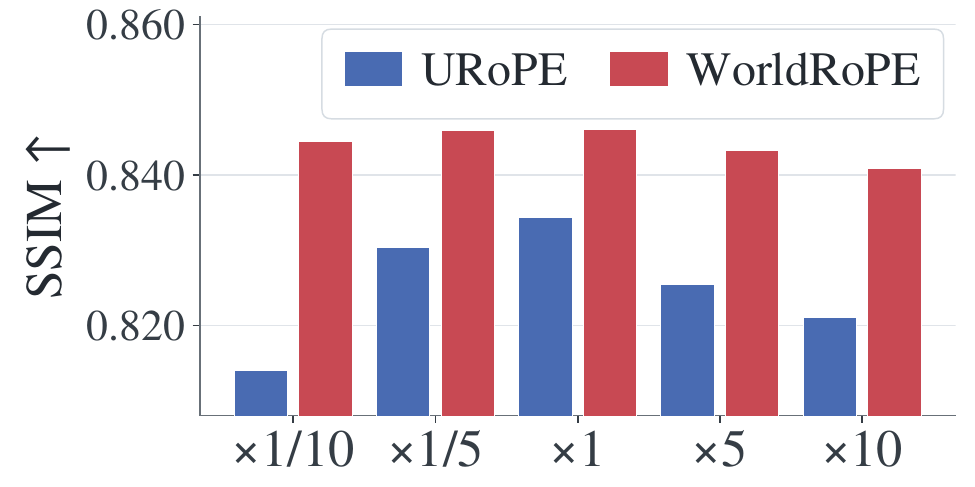}
        \caption{SSIM $\uparrow$}
        \label{fig:anchor_init_ssim}
    \end{subfigure}
    \hfill
    \begin{subfigure}[b]{0.32\linewidth}
        \centering
        \includegraphics[width=\linewidth]
            {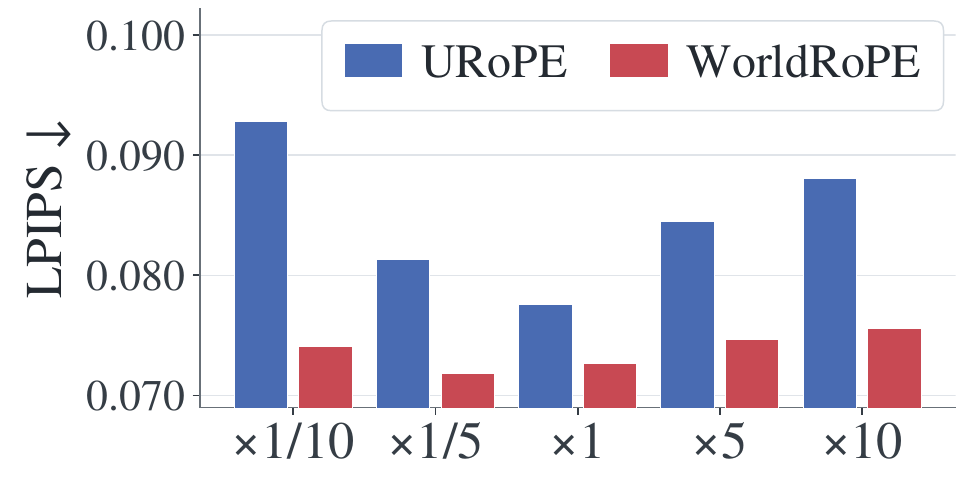}
        \caption{LPIPS $\downarrow$}
        \label{fig:anchor_init_lpips}
    \end{subfigure}
    \caption{
        Sensitivity to initialization scale on RE10K.
        The scale multiplies the initial anchors of URoPE and the initial
        bias of WorldRoPE, with $\times1$ denoting the default initialization.
        WorldRoPE remains comparatively stable across a broad range of
        initialization scales.
    }
    \label{fig:anchor_init_robustness}
\end{figure}

Figure~\ref{fig:anchor_init_robustness} examines sensitivity to the depth configuration.
We scale the predefined depth anchors of URoPE and the initial depth biases of \ourmethod{} from $0.1$ to $10$ times their default values.
\ourmethod{} maintains higher PSNR and SSIM and lower LPIPS at every tested scale, with smaller performance variations.
In particular, reducing the scale to $0.1$ substantially degrades URoPE, while \ourmethod{} remains close to its default performance.
These results indicate that the learned depth hypotheses are less sensitive to their initial placement.

\clearpage

\subsection{Visualization of world-space Gaussian primitives}
\begin{figure*}[h]
\centering
\begin{tikzpicture}
\def\qualTopGap{2pt}
\def\qualSideGap{2pt}

\node[anchor=south west, inner sep=0] (img)
    at (0.07\linewidth, 0)
    {\includegraphics[width=0.93\linewidth]
    {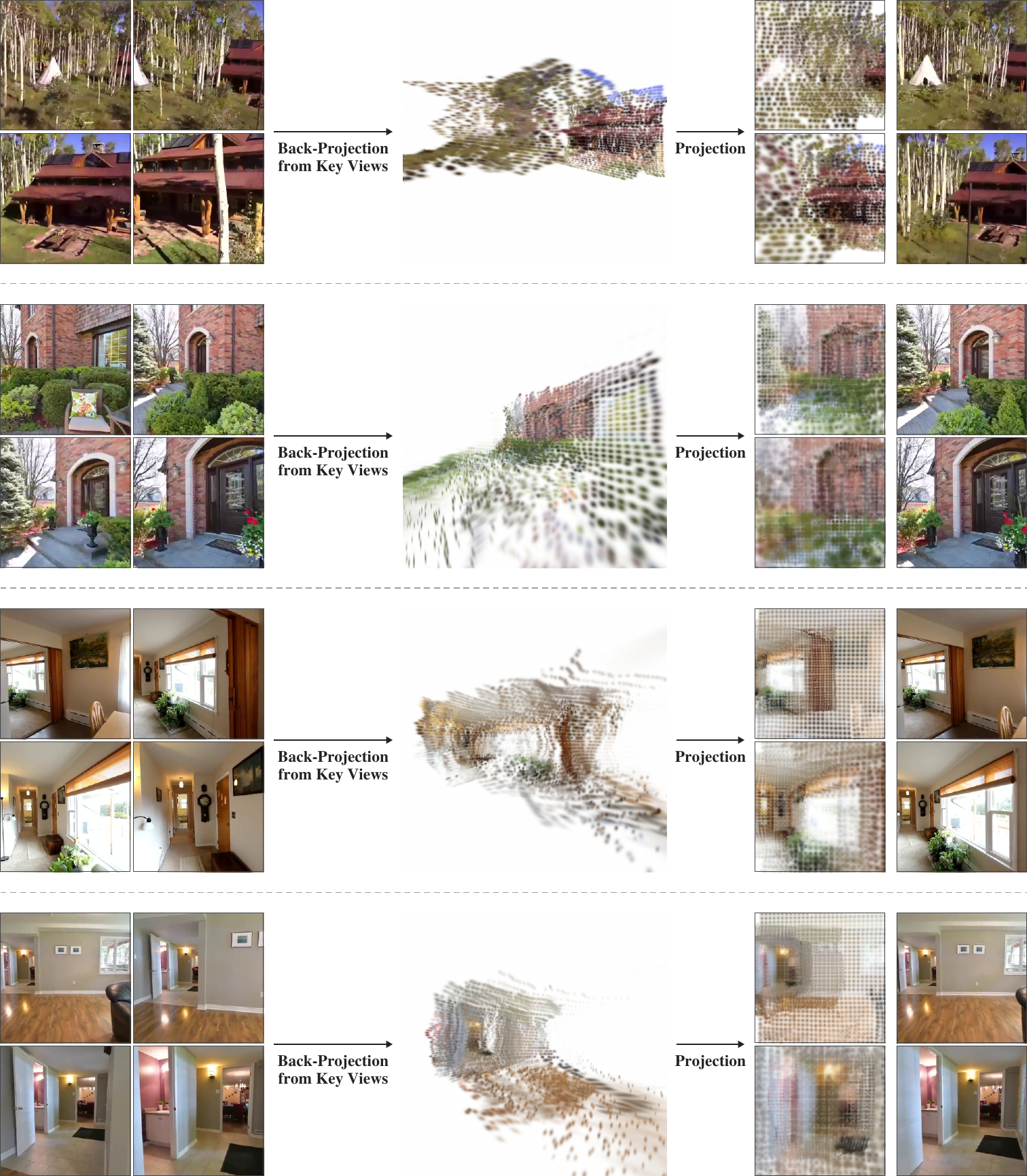}};

\foreach \x/\name in {
    0.12831/{Reference Views},
    0.52050/{Probabilistic\\World-Space Primitives},
    0.79828/{Projected Query\\Views},
    0.93651/{GT Views}
} {
    \node[anchor=south, align=center, font=\scriptsize\bfseries]
        at ($(img.north west)!\x!(img.north east)
            +(0,\qualTopGap)$)
        {\strut\name};
}

\foreach \y/\name in {
    0.88799/{(a)},
    0.62933/{(b)},
    0.37067/{(c)},
    0.11201/{(d)}
} {
    \node[anchor=east, font=\footnotesize\bfseries]
        at ($(img.south west)!\y!(img.north west)
            +(-\qualSideGap,0)$)
        {\name};
}
\end{tikzpicture}
\caption{
Visualization of the probabilistic world-space primitive projection process.
For each example, tokens from the key views are unprojected into probabilistic
world-space primitives using the predicted depth and uncertainty.
Each primitive uses the mean RGB of its source patch as color and its mixture
weight as opacity for visualization.
The primitives are then projected into the query views and compared with the
corresponding ground-truth views.
}
\label{fig:3d_gaussian_projection}
\end{figure*}

Figure~\ref{fig:3d_gaussian_projection} visualizes the predicted primitives and their projections into the query views.
The projected supports occupy geometrically plausible regions, allowing WorldRoPE to produce stronger rotary responses for key and query tokens that observe the same 3D region across views.

\clearpage
\subsection{Layer-wise geometry predictions}
\label{sec:supp_additional_layer_analysis}

\begin{figure*}[h]
    \centering
    \includegraphics[
        width=0.62\linewidth,
        height=0.76\textheight,
        keepaspectratio
    ]{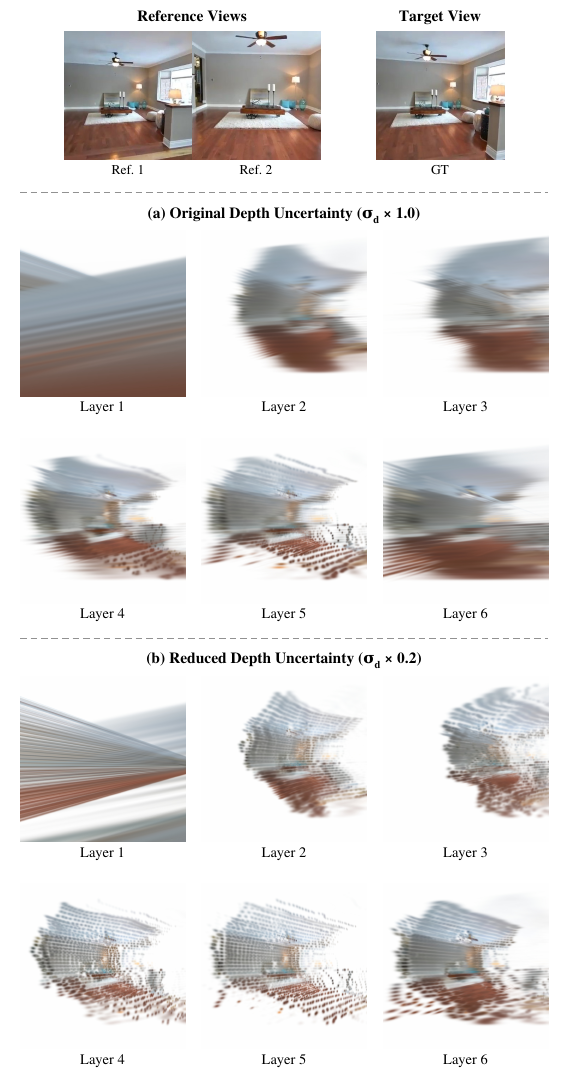}
    \vspace{-2mm}
    \caption{
    Layer-wise visualization of predicted world-space Gaussian primitives.
    We render the Gaussian primitives predicted at Layers 1--6 using the original depth uncertainty
    ($\sigma_d\times1.0$) and a reduced uncertainty
    ($\sigma_d\times0.2$).
    For visualization, each primitive is colored by the mean RGB of its
    corresponding image token, with its mixture weight used as opacity.
    }
    \label{fig:layerwise_3d_gaussians}
\end{figure*}

Figure~\ref{fig:layerwise_3d_gaussians} visualizes the world-space primitives predicted across successive Transformer layers.
Scaling $\sigma_d$ by $0.2$ yields a clearer spatial structure than using the predicted uncertainty scale directly.
Although rotary integration and explicit visualization favor different uncertainty scales, a simple rescaling of $\sigma_d$ produces a coherent spatial structure.

The predicted geometry becomes progressively more coherent through the intermediate layers, while the final layer shows a distinct pattern, possibly reflecting task-specific specialization near the output head.
Unless otherwise specified, we use the Layer~4 primitives with $\sigma_d \times 0.2$ for visualization throughout the paper.

\clearpage
\subsection{Sensitivity to camera variations}
\label{sec:additional_camera_analysis}

\begin{figure}[h]
    \centering
    \captionsetup[subfigure]{
        font=small,
        labelfont=normalfont,
        justification=centering,
        skip=-0.5mm
    }

    \begin{subfigure}[t]{0.318\textwidth}
        \centering
        \makebox[\linewidth][l]{%
            \hspace*{-0.09\linewidth}%
            \includegraphics[width=\linewidth]{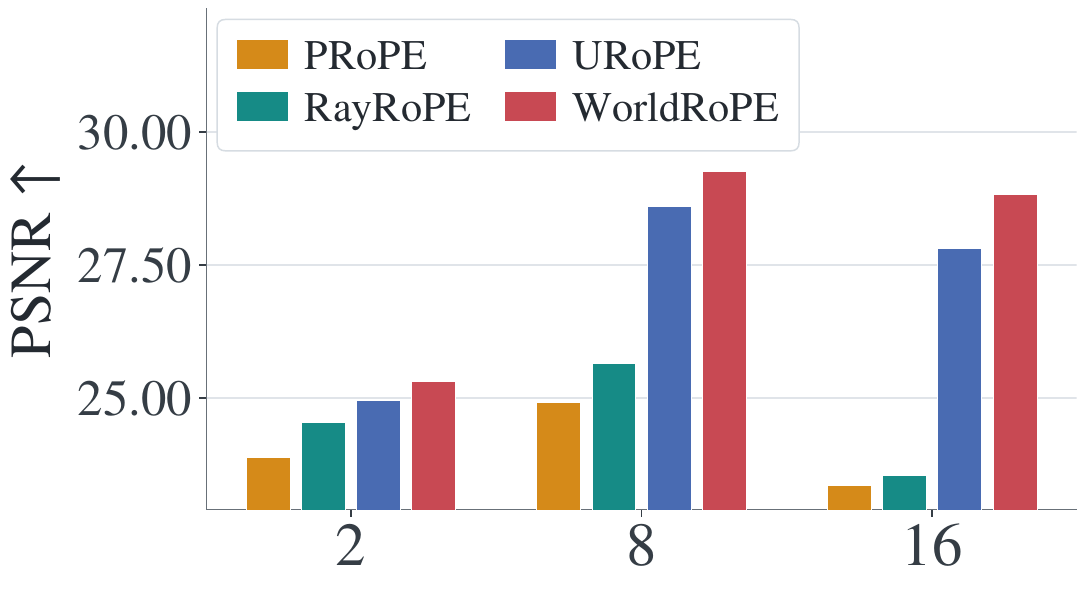}%
        }
        \caption{Number of reference views}
        \label{fig:ood_multiview_psnr}
    \end{subfigure}\hfill
    \begin{subfigure}[t]{0.318\textwidth}
        \centering
        \makebox[\linewidth][l]{%
            \hspace*{-0.09\linewidth}%
            \includegraphics[width=\linewidth]{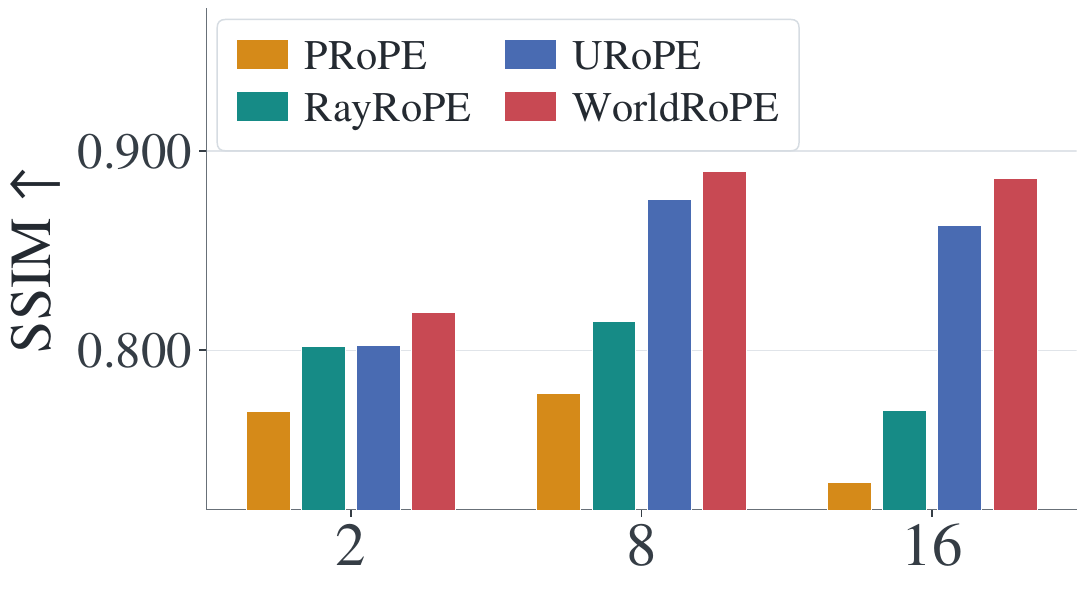}%
        }
        \caption{Number of reference views}
        \label{fig:ood_multiview_ssim}
    \end{subfigure}\hfill
    \begin{subfigure}[t]{0.318\textwidth}
        \centering
        \makebox[\linewidth][l]{%
            \hspace*{-0.09\linewidth}%
            \includegraphics[width=\linewidth]{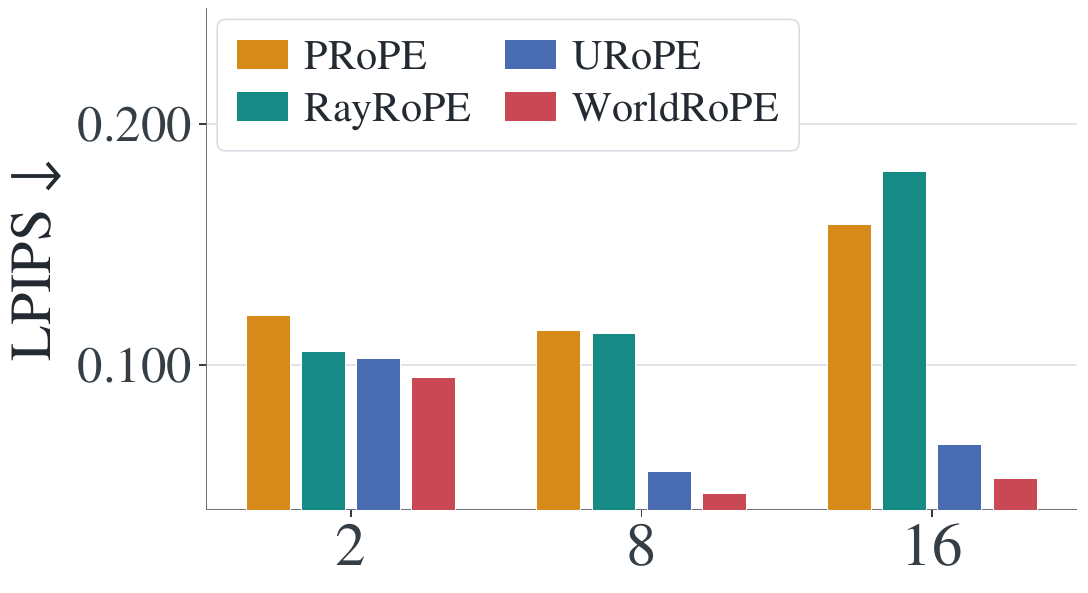}%
        }
        \caption{Number of reference views}
        \label{fig:ood_multiview_lpips}
    \end{subfigure}

    \vspace{4.8mm}

    \begin{subfigure}[t]{0.318\textwidth}
        \centering
        \makebox[\linewidth][l]{%
            \hspace*{-0.09\linewidth}%
            \includegraphics[width=\linewidth]{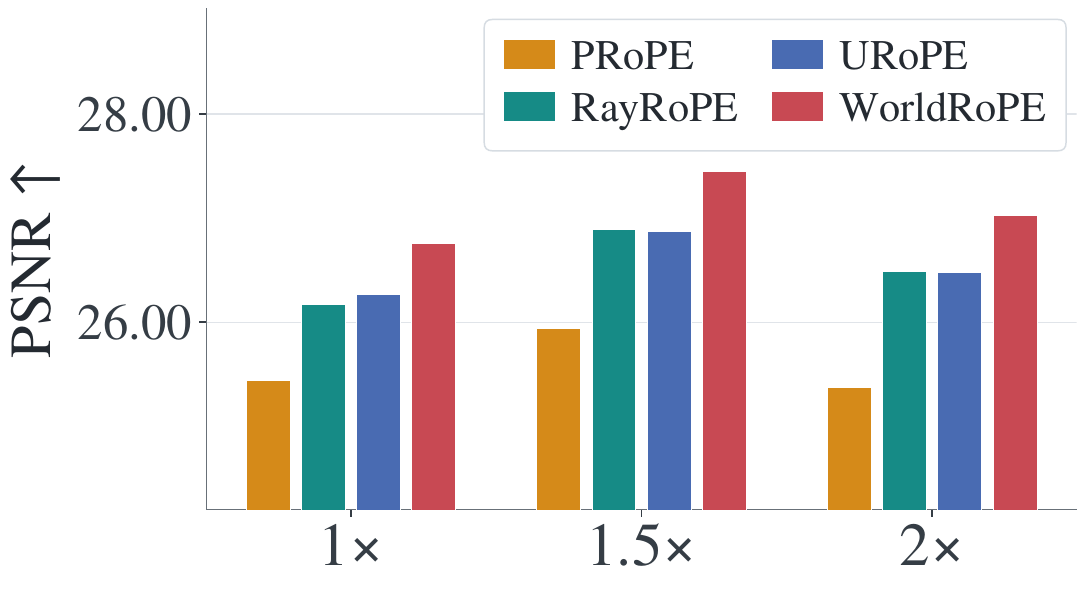}%
        }
        \caption{Zoom-in}
        \label{fig:ood_zoom_psnr}
    \end{subfigure}\hfill
    \begin{subfigure}[t]{0.318\textwidth}
        \centering
        \makebox[\linewidth][l]{%
            \hspace*{-0.09\linewidth}%
            \includegraphics[width=\linewidth]{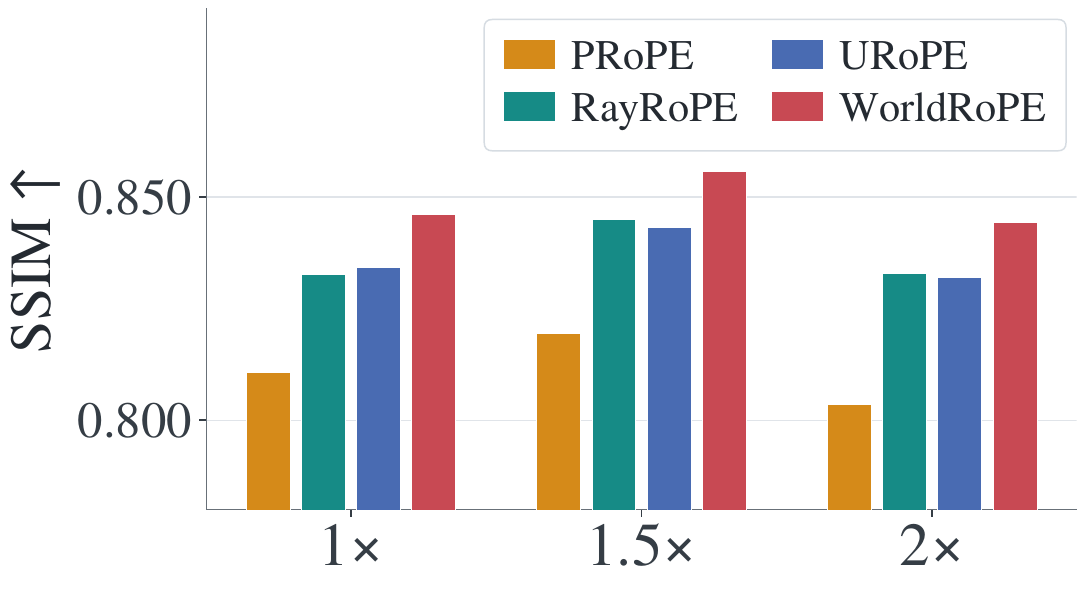}%
        }
        \caption{Zoom-in}
        \label{fig:ood_zoom_ssim}
    \end{subfigure}\hfill
    \begin{subfigure}[t]{0.318\textwidth}
        \centering
        \makebox[\linewidth][l]{%
            \hspace*{-0.09\linewidth}%
            \includegraphics[width=\linewidth]{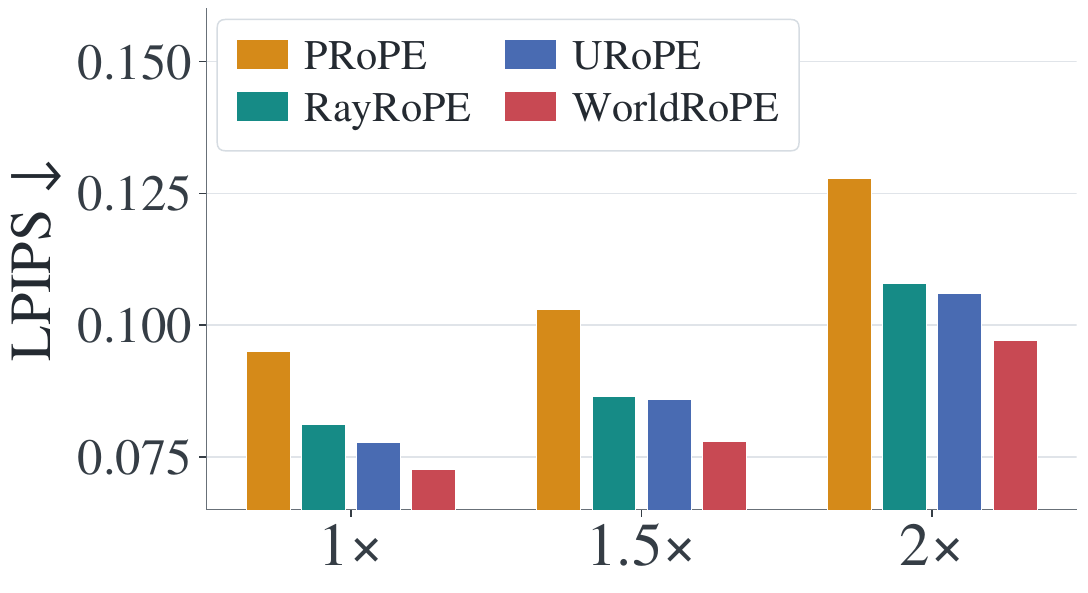}%
        }
        \caption{Zoom-in}
        \label{fig:ood_zoom_lpips}
    \end{subfigure}

    \vspace{4.8mm}

    \begin{subfigure}[t]{0.318\textwidth}
        \centering
        \makebox[\linewidth][l]{%
            \hspace*{-0.09\linewidth}%
            \includegraphics[width=\linewidth]{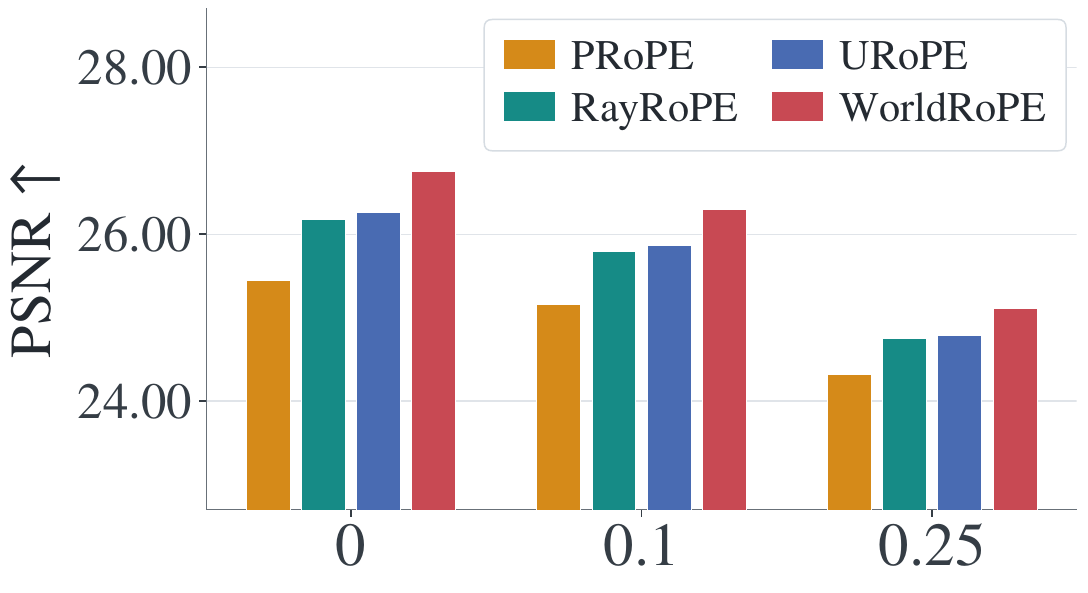}%
        }
        \caption{Rotation noise $\sigma$ ($^\circ$)}
        \label{fig:ood_rotation_psnr}
    \end{subfigure}\hfill
    \begin{subfigure}[t]{0.318\textwidth}
        \centering
        \makebox[\linewidth][l]{%
            \hspace*{-0.09\linewidth}%
            \includegraphics[width=\linewidth]{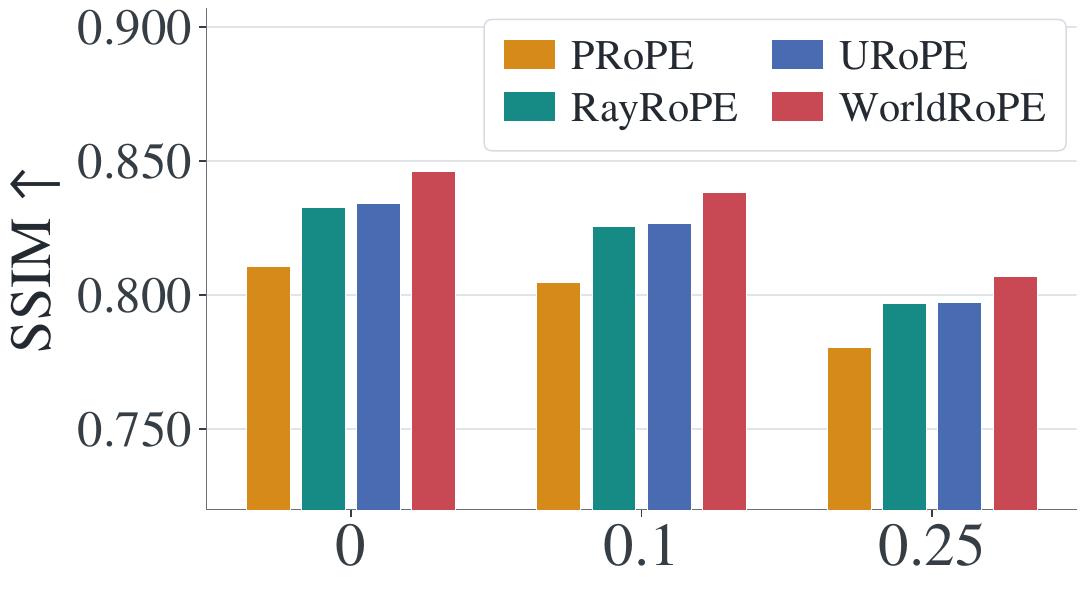}%
        }
        \caption{Rotation noise $\sigma$ ($^\circ$)}
        \label{fig:ood_rotation_ssim}
    \end{subfigure}\hfill
    \begin{subfigure}[t]{0.318\textwidth}
        \centering
        \makebox[\linewidth][l]{%
            \hspace*{-0.09\linewidth}%
            \includegraphics[width=\linewidth]{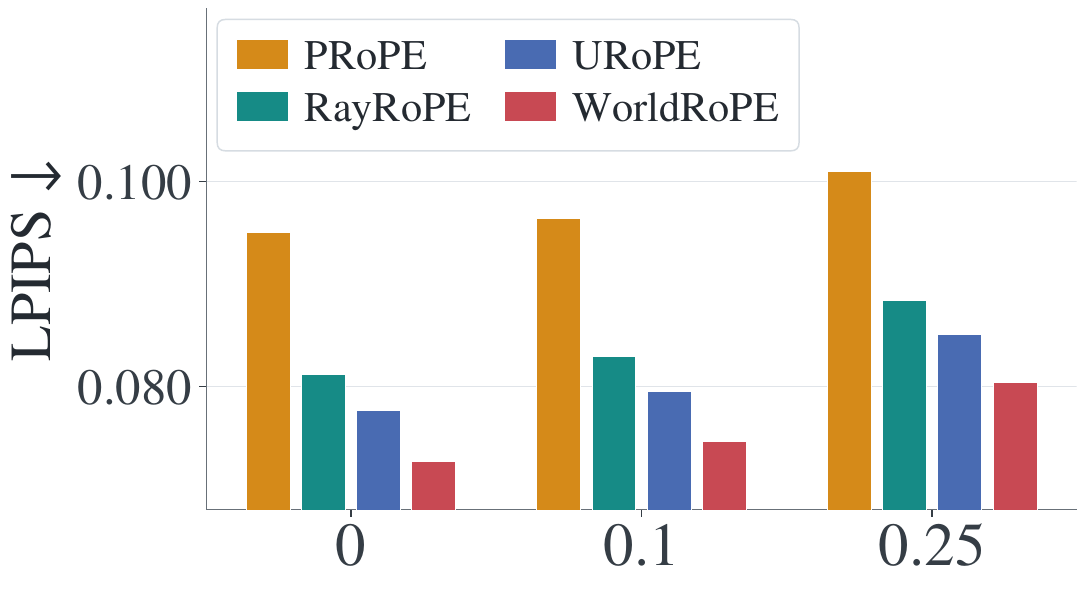}%
        }
        \caption{Rotation noise $\sigma$ ($^\circ$)}
        \label{fig:ood_rotation_lpips}
    \end{subfigure}

    \vspace{4.8mm}

    \begin{subfigure}[t]{0.318\textwidth}
        \centering
        \makebox[\linewidth][l]{%
            \hspace*{-0.09\linewidth}%
            \includegraphics[width=\linewidth]{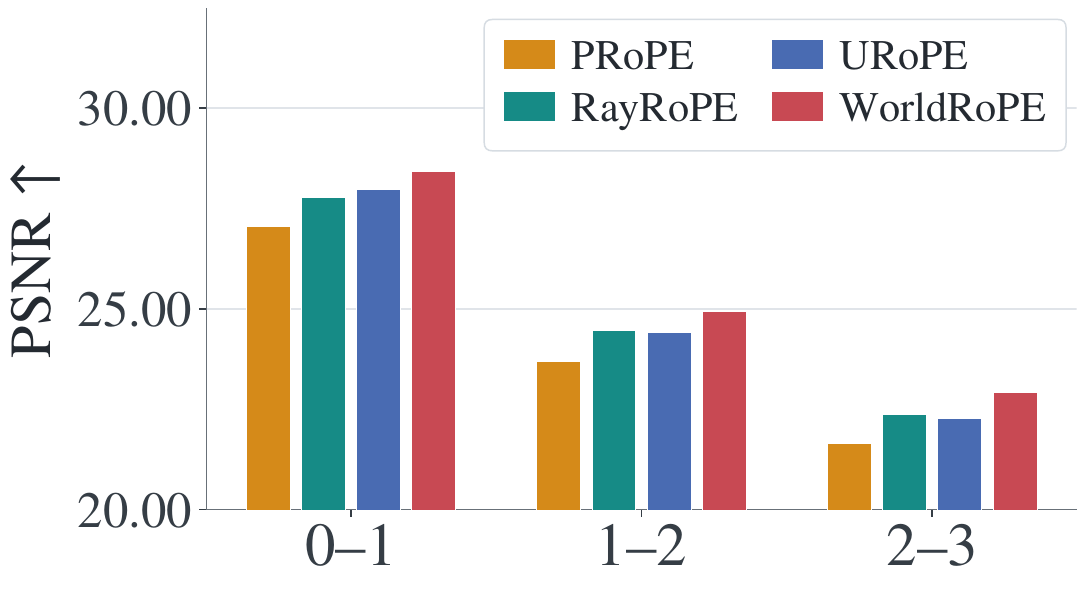}%
        }
        \caption{Reference distance interval}
        \label{fig:ood_distance_psnr}
    \end{subfigure}\hfill
    \begin{subfigure}[t]{0.318\textwidth}
        \centering
        \makebox[\linewidth][l]{%
            \hspace*{-0.09\linewidth}%
            \includegraphics[width=\linewidth]{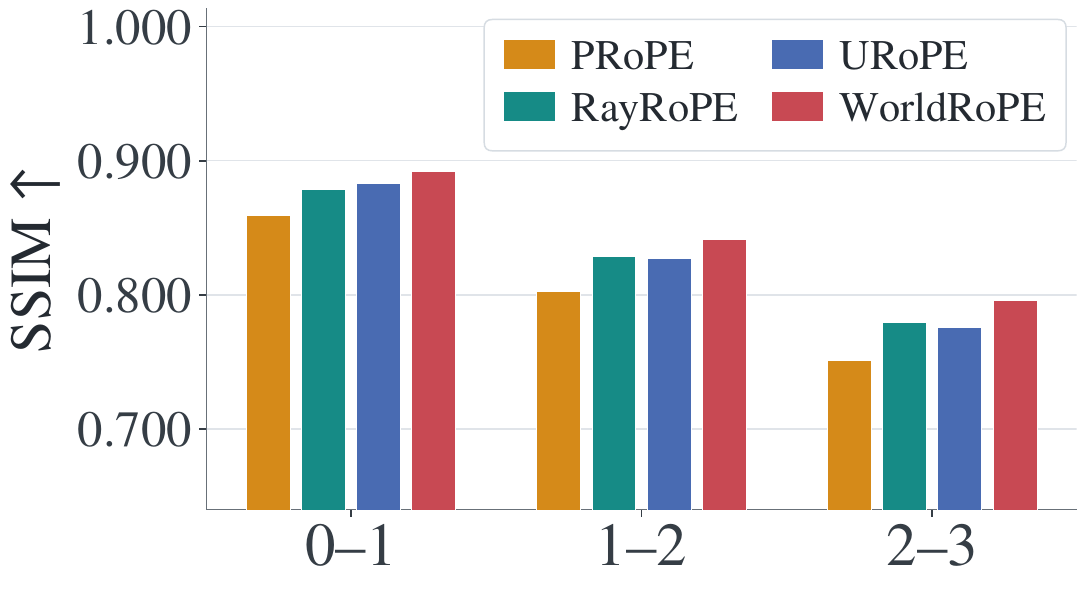}%
        }
        \caption{Reference distance interval}
        \label{fig:ood_distance_ssim}
    \end{subfigure}\hfill
    \begin{subfigure}[t]{0.318\textwidth}
        \centering
        \makebox[\linewidth][l]{%
            \hspace*{-0.09\linewidth}%
            \includegraphics[width=\linewidth]{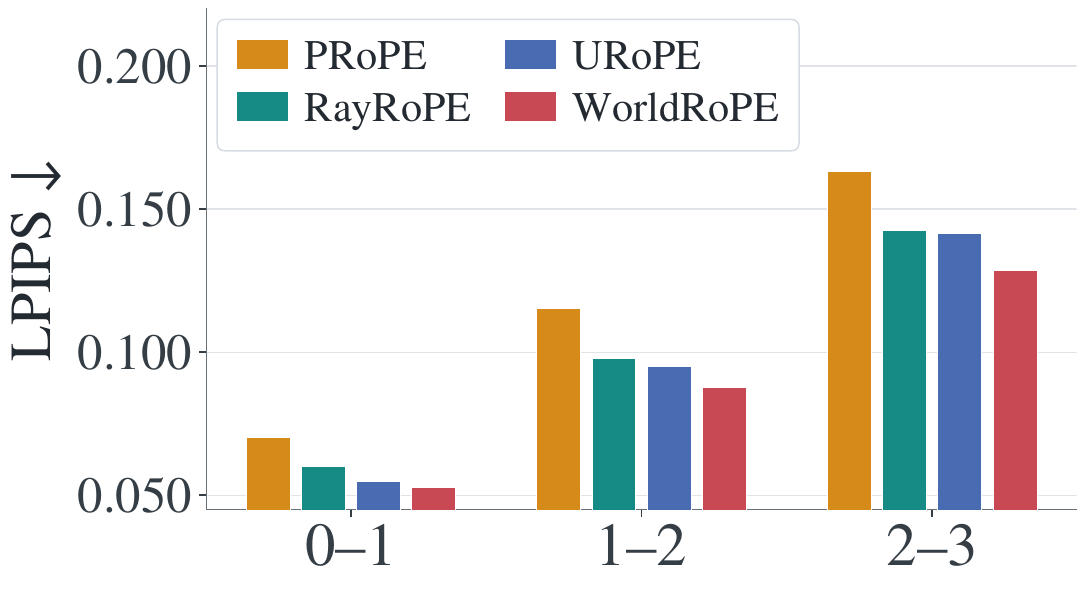}%
        }
        \caption{Reference distance interval}
        \label{fig:ood_distance_lpips}
    \end{subfigure}

    \caption{
        Robustness evaluation on RE10K under varying multi-view context and
        camera conditions. (a)--(c) vary the number of reference views at
        inference time. (d)--(f) evaluate image zoom-in. (g)--(i) evaluate
        reference-camera rotation noise. (j)--(l) evaluate reference-camera
        distance. Higher PSNR and SSIM and lower LPIPS indicate better
        performance.
    }
    \label{fig:re10k_geometric_robustness}
\end{figure}

Figure~\ref{fig:re10k_geometric_robustness} evaluates robustness under varying multi-view context and camera conditions.
In (a)--(c), although all models are trained with two reference views, WorldRoPE remains stable as the number of input views increases and achieves the strongest performance under multi-view inference.
In contrast, PRoPE and RayRoPE degrade substantially with larger view counts.
Their default implementations apply inter-camera distance normalization during training, which may contribute to this degradation, whereas URoPE and WorldRoPE do not use this normalization.

In (d)--(f), WorldRoPE maintains strong performance under image zoom-in, and similar robustness is observed under reference-camera rotation noise in (g)--(i).
WorldRoPE also maintains the strongest performance across reference-camera distance ranges in (j)--(l).
Overall, WorldRoPE remains robust across variations in multi-view context and camera geometry.

\clearpage

\subsection{Token-Adaptive Temperature}

\begin{wraptable}{r}{0.42\linewidth}
    \centering
    \vspace{-4mm}
    \caption{Ablation of token-adaptive temperature on RE10K.}
    \label{tab:temperature_ablation}
    \vspace{-2mm}
    \scalebox{0.78}{%
        \setlength{\tabcolsep}{1.5pt}
        \begin{tabular}{l|ccc}
            \toprule
            \textbf{Method}
            & \textbf{PSNR} $\uparrow$
            & \textbf{SSIM} $\uparrow$
            & \textbf{LPIPS} $\downarrow$ \\
            \midrule
            w/o Temperature
            & 26.30 & 0.835 & 0.082 \\
            w/ Learned Temperature
            & 26.33 & 0.836 & 0.082 \\
            \midrule
            \textbf{\ourmethod{}}
            & \textbf{26.75} & \textbf{0.846} & \textbf{0.073} \\
            \bottomrule
        \end{tabular}%
    }
    \vspace{-4mm}
\end{wraptable}

We examine whether the token-adaptive temperature in \ourmethod{} can be replaced by a learned scalar without modeling positional covariance.
Each key token $i$ predicts depth hypotheses $\mu_{k,i}^m$ and mixture weights $\alpha_{k,i}^m$ as in WorldRoPE, but does not model depth uncertainty $\sigma_{k,i}^m$ or patch extent, which determine the temperature.
Instead, it predicts a temperature $\tau_{k,i}^m$ for each primitive,
$\{(\alpha_{k,i}^m,\mu_{k,i}^m,\tau_{k,i}^m)\}_{m=1}^{M}=\Phi'(\mathbf{f}_{k,i})$.
As a result, the learned temperature is shared across all rotary blocks, whereas WorldRoPE derives a separate temperature from the predicted covariance for each rotary frequency. Table~\ref{tab:temperature_ablation} shows that the geometry-aware token-adaptive temperature in \ourmethod{} provides a clear performance gain over a learned token-adaptive temperature.

\subsection{Impact of Rotary Embedding on Value and Output}
\label{app:vo_output}
\begin{wraptable}{r}{0.42\linewidth}
    \centering
    \vspace{-4mm}
    \caption{Ablation of value and output rotations on RE10K.}
    \label{tab:vo_rotation_ablation}
    \vspace{-2mm}
    \scalebox{0.78}{%
        \setlength{\tabcolsep}{5.5pt}
        \begin{tabular}{l|ccc}
            \toprule
            \textbf{Method}
            & \textbf{PSNR} $\uparrow$
            & \textbf{SSIM} $\uparrow$
            & \textbf{LPIPS} $\downarrow$ \\
            \midrule
            w/o V/O rotation
             & 26.39 & 0.838 & 0.079 \\
            \midrule
            \ourmethod{}
            & \textbf{26.75} & \textbf{0.846} & \textbf{0.073} \\
            \bottomrule
        \end{tabular}%
    }
    \vspace{-4mm}
\end{wraptable}
Following GTA~\citep{miyato2024gta}, we also apply positional transformations to values and outputs. Given the attention weights $A_{ji}^{q\leftarrow k}$ from Equation~\eqref{eq:logit},
let $\mathbf{v}_i$ denote the value of key token $i$ and $\mathbf{o}_j$ the attention output for query token $j$.
The value and output transformations are then given by
\begin{equation}
\mathbf{o}_j
=
\mathcal{R}(\mathbf{u}_{q,j})^{\top}
\sum_{k,i}
A_{ji}^{q\leftarrow k}\,
\overline{\mathcal{R}}_{k,i\to q}\,
\mathbf{v}_i,
\label{eq:vo_rotation}
\end{equation}
where the factor $\mathcal{R}(\mathbf{u}_{q,j})^\top$ corresponds to the \emph{Inverse Rot.} block in Figure~\ref{fig:multi_view_att_pipeline}.
Table~\ref{tab:vo_rotation_ablation} quantifies the effect of the value and output rotations.

\subsection{Computational Cost}

WorldRoPE adds only 55K trainable parameters ($+0.23\%$) to the
25M LVSM model. These additional parameters are introduced by layer-wise
linear heads that predict depth means, depth standard deviations, and
mixture weights.

\begin{wrapfigure}{r}{0.60\linewidth}
    \centering
    \vspace{-4mm}

    \begin{subfigure}[b]{0.483\linewidth}
        \centering
        \includegraphics[width=\linewidth]
            {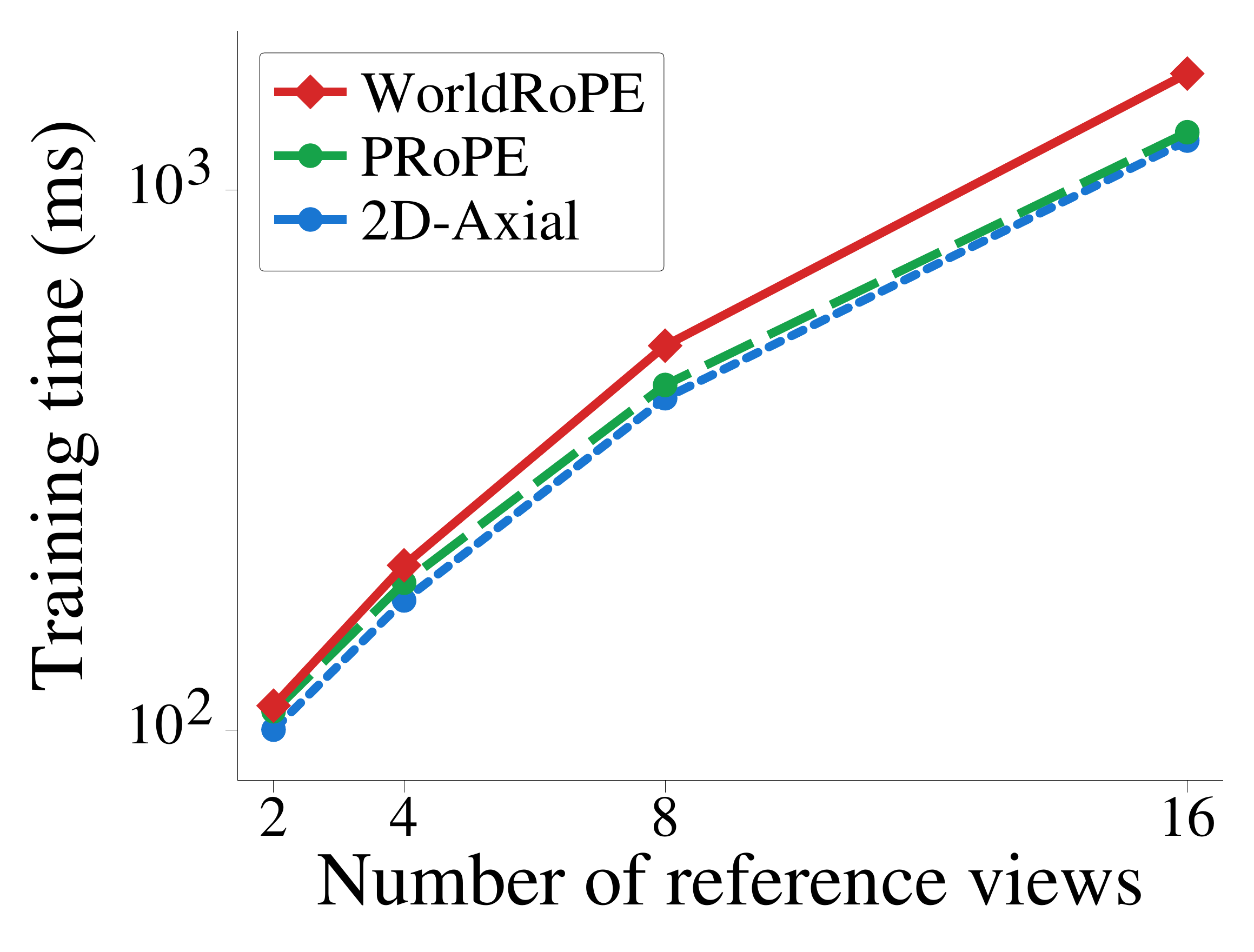}
        \caption{Training time}
        \label{fig:training_time}
    \end{subfigure}
    \hfill
    \begin{subfigure}[b]{0.483\linewidth}
        \centering
        \includegraphics[width=\linewidth]
            {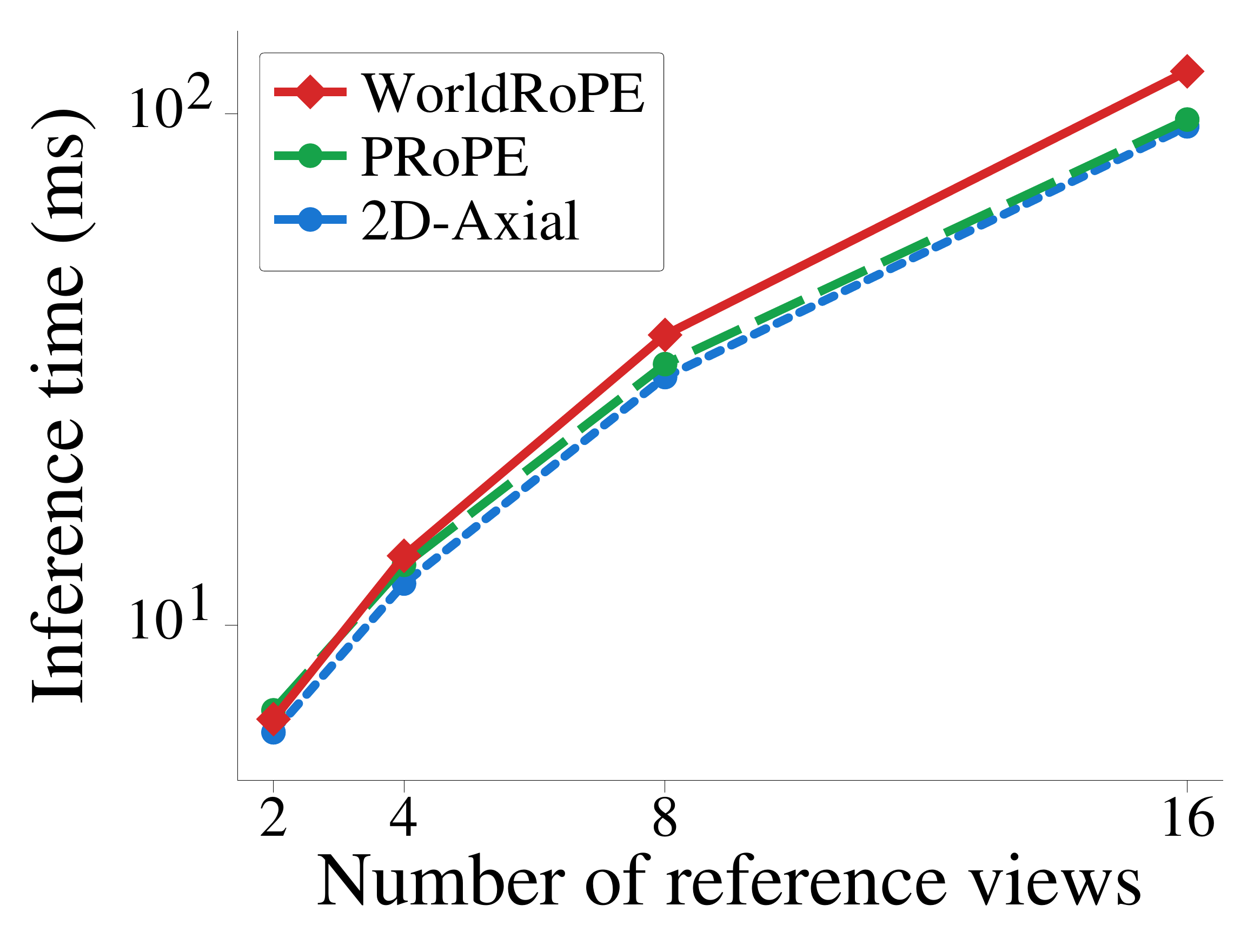}
        \caption{Inference time}
        \label{fig:inference_time}
    \end{subfigure}

    \caption{Training and inference time as the number of reference views increases.
    2D-Axial denotes standard 2D image-space RoPE.}
    \label{fig:runtime_scaling}
    \vspace{-4mm}
\end{wrapfigure}

We measure training and inference time on a single NVIDIA A6000 GPU.
Training uses a batch size of 4, while inference is measured with one
target view. All models are benchmarked with the full model compiled
using \texttt{torch.compile}.

Figure~\ref{fig:runtime_scaling} shows that the runtime of \ourmethod{} grows faster than that of 2D-Axial and PRoPE as the number of reference views increases.
To integrate with existing image-token attention as a plug-in, we project the predicted geometry into each query view at every layer, resulting in $\mathcal{O}(N^2)$ geometric computation for $N$ views.
The expected rotary operator also has a closed form for Gaussian positions in 3D.
Future models that operate directly in shared 3D space could therefore use the same formulation without repeated view-pair projections.

\clearpage
\section{Derivations and Proofs}
\label{sec:appendix_derivation}
\label{sec:scope}
We follow the notation of the main paper.
Consider a patch $j$ in a query view $q$ and a patch $i$ in a key view $k$.
Their coordinates are
$\mathbf{u}_{q,j}=(u_{q,j},v_{q,j})^\top$ and
$\mathbf{u}_{k,i}=(u_{k,i},v_{k,i})^\top$.
For each view $r$, let $\mathbf{K}_r$ and $(\mathbf{R}_r,\mathbf{t}_r)$ denote its intrinsic and extrinsic camera parameters.
The query coordinate $\mathbf{u}_{q,j}$ is deterministic, while $p_{k,i\to q}$ describes the projected positional distribution of key token $i$.
We fix one rotary frequency $\bm{\omega}$ and omit its block index unless assembling the full operator.
For generic single-primitive derivations, $\boldsymbol{\mu}$ and $\bm{\Sigma}$ denote the projected center and covariance of a distribution in $\mathbb{R}^2$.

\subsection{Expected Rotary Operators}
\label{sec:gaussian_expectation}

\paragraph{Closed-form expectation.}
For a single projected Gaussian primitive, let
$\mathbf{u}\sim\Normal(\boldsymbol{\mu},\bm{\Sigma})$ in $\mathbb{R}^2$,
suppressing token and view indices.
Since each rotary block consists of sine and cosine terms, its expectation is
\begin{equation}
\E[\RotTwo(\bm{\omega}^\top\mathbf{u})]
=
\begin{bmatrix}
\E[\cos(\bm{\omega}^\top\mathbf{u})]
&
-\E[\sin(\bm{\omega}^\top\mathbf{u})]
\\
\E[\sin(\bm{\omega}^\top\mathbf{u})]
&
\E[\cos(\bm{\omega}^\top\mathbf{u})]
\end{bmatrix}.
\label{eq:expected_rotation_entries}
\end{equation}
We compute these terms together using Euler's identity,
$e^{\mathrm{i}\theta}=\cos\theta+\mathrm{i}\sin\theta$.
The characteristic function of the Gaussian gives
\begin{equation}
\begin{aligned}
\E[e^{\mathrm{i}\bm{\omega}^\top\mathbf{u}}]
&=
\int_{\mathbb{R}^2}
e^{\mathrm{i}\bm{\omega}^\top\mathbf{u}}
\Normal(\mathbf{u};\boldsymbol{\mu},\bm{\Sigma})
\,d\mathbf{u}
\\
&=
\exp\!\left(
-\tfrac12\bm{\omega}^\top\bm{\Sigma}\bm{\omega}
\right)
e^{\mathrm{i}\bm{\omega}^\top\boldsymbol{\mu}}.
\end{aligned}
\label{eq:characteristic}
\end{equation}
Taking the real and imaginary parts yields
\begin{equation}
\begin{aligned}
\E[\cos(\bm{\omega}^\top\mathbf{u})]
&=a\cos(\bm{\omega}^\top\boldsymbol{\mu}),\\
\E[\sin(\bm{\omega}^\top\mathbf{u})]
&=a\sin(\bm{\omega}^\top\boldsymbol{\mu}),
\end{aligned}
\qquad
a:=\exp\!\left(
-\tfrac12\bm{\omega}^\top\bm{\Sigma}\bm{\omega}
\right).
\label{eq:expected_trigonometric_terms}
\end{equation}
Both terms share the same attenuation $a$.
Substituting them into Equation~\eqref{eq:expected_rotation_entries} therefore gives
\begin{equation}
\E[\RotTwo(\bm{\omega}^\top\mathbf{u})]
=
a\RotTwo(\bm{\omega}^\top\boldsymbol{\mu}).
\label{eq:gaussian_block_expectation}
\end{equation}
Thus, the expected block is a rotation at the projected center multiplied by covariance-dependent attenuation.
By linearity of expectation, combining the primitive contributions with weights $\alpha_{k,i}^m$ gives Equation~\eqref{eq:mixture_operator}.

\paragraph{Equivalent single-rotation form of the mixture.}
Although the rotary operator combines contributions from $M$ Gaussian components, their weighted sum can be represented exactly by a single scaled rotation for each rotary frequency block.

For a fixed rotary frequency $\bm{\omega}$, recall from Eq.~\eqref{eq:mixture_operator} that
$\overline{R}_{k,i\to q}=\sum_{m=1}^{M}\alpha_{k,i}^m a_{k,i\to q}^m
R(\bm{\omega}^\top\boldsymbol{\mu}_{k,i\to q}^m)$,
where
$a_{k,i\to q}^m=\exp(-\tfrac12\bm{\omega}^\top\bm{\Sigma}_{k,i\to q}^m\bm{\omega})$.
Let $\theta_m=\bm{\omega}^\top\boldsymbol{\mu}_{k,i\to q}^m$.
The combined cosine and sine coefficients are given by
\begin{align}
C_{k,i\to q}
&=
\sum_{m=1}^{M}\alpha_{k,i}^m a_{k,i\to q}^m\cos\theta_m,
&
S_{k,i\to q}
&=
\sum_{m=1}^{M}\alpha_{k,i}^m a_{k,i\to q}^m\sin\theta_m.
\end{align}
The mixture operator can then be rewritten as follows:
\begin{align}
\overline{R}_{k,i\to q}
&=
\begin{bmatrix}
C_{k,i\to q} & -S_{k,i\to q}\\
S_{k,i\to q} & C_{k,i\to q}
\end{bmatrix}
=
\bar a_{k,i\to q}R(\bar\theta_{k,i\to q}).
\end{align}
The effective attenuation and phase are given by
\begin{align}
\bar a_{k,i\to q}
&=
\sqrt{C_{k,i\to q}^{2}+S_{k,i\to q}^{2}},
&
\bar\theta_{k,i\to q}
&=
\operatorname{atan2}(S_{k,i\to q},C_{k,i\to q}).
\end{align}
When $\bar a_{k,i\to q}=0$, the operator is zero and the phase is arbitrary.
Thus, for each rotary frequency block, the Gaussian mixture is exactly equivalent to a single scaled rotation with effective attenuation $\bar a_{k,i\to q}$ and phase $\bar\theta_{k,i\to q}$.

\paragraph{Extension to probabilistic queries and keys.}
\ourmethod{} naturally generalizes to the case where both query and key positions are probabilistic.
To describe their rotary interaction, consider independent positions $\mathbf{u}\sim\Normal(\boldsymbol{\mu},\bm{\Sigma})$ and $\mathbf{u}'\sim\Normal(\boldsymbol{\mu}',\bm{\Sigma}')$ in a common coordinate system.
Their relative position is given by
\begin{align}
\mathbf{u}-\mathbf{u}'
\sim\Normal\!\left(
\boldsymbol{\mu}-\boldsymbol{\mu}',\,
\bm{\Sigma}+\bm{\Sigma}'
\right),
\end{align}
which allows us to compute its expected rotation using the same closed-form expression.
Under the independence assumption, this expectation is equivalently obtained from the individual expected operators as follows:
\begin{align}
\E[\Rot(\mathbf{u}-\mathbf{u}')]
=
\bigl(\E[\Rot(\mathbf{u}')]\bigr)^\top
\E[\Rot(\mathbf{u})].
\end{align}

\paragraph{Derivation of the sensitivity bound.}
\label{app:sensitivity_bound}
For a projected Gaussian with attenuation $a$, consider perturbing its projected center
$\boldsymbol{\mu}$ by $\boldsymbol{\delta}$.
Then,
\begin{align}
&\left\lVert
aR\!\left(
\bm{\omega}^{\top}(\boldsymbol{\mu}+\boldsymbol{\delta})
\right)
-
aR\!\left(
\bm{\omega}^{\top}\boldsymbol{\mu}
\right)
\right\rVert_2 \\
&\qquad=
a\left\lVert
R\!\left(
\bm{\omega}^{\top}\boldsymbol{\mu}
\right)
\left[
R\!\left(
\bm{\omega}^{\top}\boldsymbol{\delta}
\right)
-\mathbf{I}
\right]
\right\rVert_2 \\
&\qquad=
a\left\lVert
R\!\left(
\bm{\omega}^{\top}\boldsymbol{\delta}
\right)
-\mathbf{I}
\right\rVert_2 \\
&\qquad=
2a\left|
\sin\!\left(
\frac{\bm{\omega}^{\top}\boldsymbol{\delta}}{2}
\right)
\right|
\leq
a\left|
\bm{\omega}^{\top}\boldsymbol{\delta}
\right|.
\end{align}

\subsection{Compatibility with Lie Group Theory}
\label{sec:lie_group}
Lie-group-based rotary encodings~\citep{ostmeier2024liere, yu2025comrope} parameterize rotations in high-dimensional feature spaces through matrix exponentials; we consider $\Rot(\mathbf{u})=\exp(\mathbf{A}u+\mathbf{B}v)$,
where $\mathbf{u}=(u,v)^\top$ and $\mathbf{A},\mathbf{B}\in\mathbb{R}^{d\times d}$ are commuting skew-symmetric generators.

\ourmethod{} can be applied to these rotations by integrating over the uncertainty in positional coordinates.
For $\mathbf{u}\sim\Normal(\boldsymbol{\mu},\bm{\Sigma})$, the expected rotation matrix is given by

\begin{align}
\E[\Rot(\mathbf{u})]
&=
\int_{\mathbb{R}^2}
e^{\mathbf{A}u+\mathbf{B}v}
\Normal(\mathbf{u};\boldsymbol{\mu},\bm{\Sigma})
\,d\mathbf{u}\\
&=
e^{\mathbf{A}\mu_u+\mathbf{B}\mu_v}
\E\!\left[e^{\mathbf{A}\delta u+\mathbf{B}\delta v}\right]\\
&=
e^{\mathbf{A}\mu_u+\mathbf{B}\mu_v}
\exp\!\left(
\tfrac12\E\!\left[
(\mathbf{A}\delta u+\mathbf{B}\delta v)^2
\right]
\right)\\
&=
e^{\mathbf{A}\mu_u+\mathbf{B}\mu_v}
\exp\!\Bigl(
\tfrac12\bigl[
\Sigma_{uu}\mathbf{A}^2
+2\Sigma_{uv}\mathbf{A}\mathbf{B}
+\Sigma_{vv}\mathbf{B}^2
\bigr]
\Bigr).
\label{eq:gaussian_rotary_closed_form}
\end{align}

As in the main paper, the mean determines the rotation, while the covariance determines the attenuation due to positional uncertainty.
In particular, the term $2\Sigma_{uv}\mathbf{A}\mathbf{B}$ captures correlations between the two coordinate axes, as in mixed RoPE (Section~\ref{subsec:mixedrope}).

\paragraph{Comparison with uniform integration.}
For a one-dimensional interval as used in RayRoPE~\citep{wu2026rayrope}, consider
$\mathbf{u}=\boldsymbol{\mu}+t\mathbf{s}$,
where $t\sim\operatorname{Unif}[-1,1]$ and
$\mathbf{s}=(s_u,s_v)^\top$ denotes the interval half-extent.
With $\mathbf{H}=\mathbf{A}s_u+\mathbf{B}s_v$, uniform integration gives
\begin{equation}
\E_{\mathrm{Unif}}[\Rot(\mathbf{u})]
=
e^{\mathbf{A}\mu_u+\mathbf{B}\mu_v}
\operatorname{sinhc}(\mathbf{H}),
\end{equation}
where $\operatorname{sinhc}(\mathbf{X})=\sum_{n=0}^{\infty}\mathbf{X}^{2n}/(2n+1)!$.
The uniform expectation depends on the interval direction and extent, whereas the Gaussian expectation in Equation~\eqref{eq:gaussian_rotary_closed_form} incorporates the positional covariance $\bm{\Sigma}$, including variation along both coordinate axes and their correlation.
Within each planar rotation block, uniform integration yields a sinc factor that can change sign, while Gaussian integration yields positive exponential attenuation.
Training also requires differentiating the expected rotary operator with respect to the learned generators.
For invertible $\mathbf{H}$, the uniform factor can be written as $\mathbf{H}^{-1}\sinh(\mathbf{H})$.
Direct differentiation of this form introduces inverse-dependent terms that can be numerically sensitive near singularity.
In contrast, WorldRoPE uses the Gaussian expectation in Equation~\eqref{eq:gaussian_rotary_closed_form}, avoiding this inverse-based differentiation and reducing potential numerical sensitivity near singular configurations.

\subsection{Gaussian Primitive Construction}
\label{sec:gaussian_construction}
A finite image patch observes a spatial region whose extent cannot be represented by its central ray alone.
Mip-NeRF addresses this geometric property by approximating conical frustums with Gaussians~\citep{barron2021mip}.
We follow the same perspective of representing spatial extent with Gaussian moments, but construct our primitives from a rectangular patch footprint and predicted depth uncertainty.
Below, we derive the mean and covariance in Equation~\eqref{eq:local_cov} through first-order unprojection.

\paragraph{Patch and depth offsets.}
Let $\boldsymbol{\epsilon}\in\mathbb{R}^2$ denote the zero-mean pixel offset from the patch center, with covariance $\mathbf{C}$.
For a uniformly weighted rectangular footprint of size $\Delta_u\times\Delta_v$,
\begin{equation}
\mathbf{C}
=\diag\!\left(
\frac{\Delta_u^2}{12},
\frac{\Delta_v^2}{12}
\right).
\end{equation}
For depth hypothesis $m$, write
$D_{k,i}^m=\mu_{k,i}^m+\delta d$, where
$\E[\delta d]=0$ and
$\E[(\delta d)^2]=(\sigma_{k,i}^m)^2$.
Assuming independent pixel and depth offsets, the camera-space position is
\begin{equation}
\mathbf{X}_{k,i}^m
=(\mu_{k,i}^m+\delta d)
(\mathbf{r}_{k,i}+\mathbf{B}_k\boldsymbol{\epsilon}),
\end{equation}
where $\mathbf{r}_{k,i}=\mathbf{K}_k^{-1}(u_{k,i},v_{k,i},1)^\top$ and
$\mathbf{B}_k$ contains the first two columns of $\mathbf{K}_k^{-1}$.

\paragraph{First-order Gaussian approximation.}
Linearizing the unprojection omits the product
$\delta d\,\mathbf{B}_k\boldsymbol{\epsilon}$, yielding
\begin{equation}
\mathbf{X}_{k,i,\mathrm{lin}}^m
=
\mu_{k,i}^m\mathbf{r}_{k,i}
+\mu_{k,i}^m\mathbf{B}_k\boldsymbol{\epsilon}
+\mathbf{r}_{k,i}\delta d.
\label{eq:unprojection_first_order}
\end{equation}
We define the Gaussian primitive by matching the mean and covariance of this linearized position.
Since both offsets have zero mean, its center is
\begin{equation}
\boldsymbol{\mu}_{k,i}^m
:=\E[\mathbf{X}_{k,i,\mathrm{lin}}^m]
=\mu_{k,i}^m\mathbf{r}_{k,i}.
\end{equation}
Subtracting this mean and expanding the outer product gives
\begin{equation}
\begin{aligned}
\bm{\Sigma}_{k,i}^m
&:=\Cov(\mathbf{X}_{k,i,\mathrm{lin}}^m)\\
&=\E\!\left[
(\mu_{k,i}^m\mathbf{B}_k\boldsymbol{\epsilon}+\mathbf{r}_{k,i}\delta d)
(\mu_{k,i}^m\mathbf{B}_k\boldsymbol{\epsilon}+\mathbf{r}_{k,i}\delta d)^\top
\right]\\
&=(\mu_{k,i}^m)^2\mathbf{B}_k
\E[\boldsymbol{\epsilon}\boldsymbol{\epsilon}^\top]
\mathbf{B}_k^\top
+\E[(\delta d)^2]\mathbf{r}_{k,i}\mathbf{r}_{k,i}^\top\\
&\quad
+\mu_{k,i}^m\mathbf{B}_k\E[\boldsymbol{\epsilon}\,\delta d]\mathbf{r}_{k,i}^\top
+\mu_{k,i}^m\mathbf{r}_{k,i}\E[\delta d\,\boldsymbol{\epsilon}^\top]\mathbf{B}_k^\top.
\end{aligned}
\end{equation}
The offsets are independent, so
$\E[\boldsymbol{\epsilon}\,\delta d]
=\E[\boldsymbol{\epsilon}]\E[\delta d]=\mathbf{0}$.
The cross terms therefore vanish, and substituting the offset moments yields
\begin{equation}
\bm{\Sigma}_{k,i}^m
=
(\mu_{k,i}^m)^2\mathbf{B}_k\mathbf{C}\mathbf{B}_k^\top
+(\sigma_{k,i}^m)^2\mathbf{r}_{k,i}\mathbf{r}_{k,i}^\top.
\label{eq:unprojection_covariance_approx}
\end{equation}
This recovers Equation~\eqref{eq:local_cov}, with the two covariance terms capturing the patch footprint and depth uncertainty, respectively.

\clearpage 
\section{Implementation Details}
\label{sec:experimental_details}

\subsection{Common Implementation}
\label{sec:experimental_details_common}
Across all experiments, we implement each predictor $\Phi$ as a layer-wise linear head.
Mixture weights are normalized using a softmax, depth standard deviations are
parameterized exponentially, and depth means are clamped to be nonnegative.
These predicted mixture weights, depth means, and depth standard deviations
receive no direct auxiliary supervision and are learned end-to-end through
the task-specific objective. Table~\ref{tab:implementation-summary} summarizes the implementation settings for each task.

\subsection{Novel View Synthesis: LVSM~\citep{jin2024lvsm}}
\label{sec:experimental_details_lvsm}
Our experiments use the 25M-parameter LVSM architecture implemented in the URoPE codebase~\citep{xie2026urope}~\footnote{\url{https://github.com/Applied-Intuition-Open-Source/URoPE}}.
The model consists of six Transformer layers with a hidden dimension of 768, 16 attention heads, and a feedforward dimension of 1024.
Images have a resolution of $256\times256$ and are divided into $8\times8$ patches.
Each training sample consists of two reference views and one supervised target view.
We train the model for 80K steps on two GPUs with a global batch size of 16, using MSE loss and VGG perceptual loss with weights of 1 and 0.5, respectively.

For the positional-encoding baselines, we use the Pl\"ucker Ray, GTA, and
PRoPE implementations provided in the URoPE codebase, while URoPE is
evaluated using its native implementation. We implement RayRoPE following
its official codebase.
All methods use the same backbone and training protocol.

We train separate models on RealEstate10K (RE10K)~\citep{zhou2018stereo} and Objaverse~\citep{deitke2023objaverse}. For RE10K, we follow the train/test scene split used in the official
RayRoPE codebase~\citep{wu2026rayrope}~\footnote{\url{https://github.com/Lucas-707/RayRoPE}}. For Objaverse, we use object renderings from
eight camera directions spanning the full azimuth range.

For the mixed RoPE~\citep{heo2024ropevit}  variants, we initialize learnable frequency vectors
with the same magnitudes as the corresponding original frequencies but
with randomly sampled unit directions. For RayRoPE, Table~\ref{tab:mixed_rope_results} reports results from mixing the $u$ and $v$ coordinates without mixing inverse depth, since additionally mixing inverse depth reduced PSNR by 0.28 dB on RE10K and 1.07 dB on Objaverse.

\subsection{Depth Estimation: UniMatch~\citep{xu2023unifying}}
\label{sec:experimental_details_unimatch}

We use the official UniMatch codebase~\citep{xu2023unifying}~\footnote{\url{https://github.com/autonomousvision/unimatch}}.
Models are trained from scratch on the Scenes11~\citep{ummenhofer2017demon}, SUN3D~\citep{xiao2013sun3d}, and RGBD~\citep{sturm2012benchmark} datasets for 50K steps on two GPUs with a global
batch size of 16. The Transformer has six layers, 144 feature channels,
and three attention heads. Training uses $224\times288$ images, with
evaluation at $240\times320$. Self-attention retains the $2\times2$
shifted-window structure, while cross-attention is global. Other settings follow the UniMatch codebase.

The geometry-aware encodings of GTA, PRoPE, URoPE, RayRoPE, and
\ourmethod{} in Table~\ref{tab:unimatch_depth_results} are applied
only to cross-attention, whereas the Pl\"ucker Ray baseline adds
linearly projected Pl\"ucker coordinates to the image features. GTA, PRoPE, URoPE, and RayRoPE follow their
official implementations, adapted to UniMatch.

For evaluation, we follow the official UniMatch protocol and use the same
benchmark evaluation sets and preprocessing provided in its codebase.
For unseen-data evaluation on ScanNet++~\citep{yeshwanth2023scannet++}, we adopt the validation and test
splits defined by MapAnything~\citep{keetha2026mapanything}. To account for scale differences between the training data and unseen ScanNet++, we align each prediction to the ground-truth scale using least squares before computing Abs Rel and RMSE.

\subsection{3D Geometry Estimation: VGGT~\citep{wang2025vggt}}
\label{sec:experimental_details_vggt}
We initialize from the pretrained VGGT-1B model, which contains 24
alternating frame-wise and global self-attention stages with
1,024-dimensional features and 16 attention heads. Following the geometry-conditioning design of Pi3X~\citep{wang2025pi}, we insert eight geometry-attention blocks with the same hidden dimension as the VGGT backbone, placing one after every third global-attention block. We fully train the eight added blocks and the camera and depth prediction
heads, while applying rank-8 LoRA~\citep{hu2022lora} to the corresponding eight pretrained
global-attention blocks and keeping the remainder of the pretrained
model frozen. We train on the ScanNet++ training split provided by
MapAnything~\citep{keetha2026mapanything}, comprising 896 scenes.
We evaluate on the corresponding 30-scene ScanNet++ test split.

\begin{table*}[h]
\centering
\captionsetup{skip=4pt}
\caption{
Summary of the task-specific implementations of \ourmethod{}.
}
\label{tab:implementation-summary}

\begingroup
\fontsize{10}{11.5}\selectfont
\setlength{\tabcolsep}{5pt}
\renewcommand{\arraystretch}{1.18}

\setlength{\heavyrulewidth}{0.8pt}
\setlength{\lightrulewidth}{0.4pt}
\setlength{\aboverulesep}{3pt}
\setlength{\belowrulesep}{3pt}

\begin{tabular}{@{}
    >{\centering\arraybackslash}m{0.20\linewidth}
    *{3}{>{\centering\arraybackslash}
        m{\dimexpr(0.80\linewidth-6\tabcolsep)/3\relax}}
@{}}
\toprule

\textbf{Setting}
& \shortstack[c]{
    \strut\textbf{Novel-view}\\
    \strut\textbf{synthesis}
}
& \shortstack[c]{
    \strut\textbf{Depth}\\
    \strut\textbf{estimation}
}
& \shortstack[c]{
    \strut\textbf{3D geometry}\\
    \strut\textbf{estimation}
}
\\
\midrule

Backbone
& LVSM-25M
& UniMatch
& VGGT-1B
\\
\addlinespace[3pt]

Training data
& RE10K or Objaverse
& \shortstack[c]{
    \strut Scenes11, SUN3D,\\
    \strut RGB-D SLAM
}
& \shortstack[c]{
    \strut ScanNet++\\
    \strut (896 scenes)
}
\\
\addlinespace[3pt]

Evaluation data
& \shortstack[c]{
    \strut RE10K or Objaverse\\
    \strut test split
}
& \shortstack[c]{
    \strut Scenes11, SUN3D,\\
    \strut RGB-D SLAM,\\
    \strut and ScanNet++
}
& \shortstack[c]{
    \strut ScanNet++\\
    \strut (30 held-out scenes)
}
\\
\midrule

Training views
& \shortstack[c]{
    \strut 2 references\\
    \strut + 1 target
}
& 2 views
& 4 views
\\
\addlinespace[2pt]

Resolution (train)
& $256\times256$
& $224\times288$
& $518\times294$
\\

Resolution (eval.)
& $256\times256$
& $240\times320$
& $518\times294$
\\

Components ($M$)
& 4
& 3
& 4
\\
\midrule

Training objective
& $\mathcal{L}_{\mathrm{MSE}}
   + 0.5\mathcal{L}_{\mathrm{VGG}}$
& $20\mathcal{L}_{\mathrm{inv}}
   + 20\mathcal{L}_{\nabla\mathrm{inv}}$
& \shortstack[c]{
    \strut Confidence-weighted\\
    \strut robust log-depth and\\
    \strut world-point regression
}
\\
\addlinespace[3pt]

Training steps
& 80K & 50K & 50K
\\

GPUs
& 2 & 2 & 1
\\

Batch size
& 16 & 16 & 2
\\

Training precision
& FP16 & BF16 & BF16
\\

\bottomrule
\end{tabular}
\par
\endgroup
\end{table*}

\clearpage
\section{Additional Qualitative Results}
\label{sec:additional_qualitative}

\subsection{Novel view synthesis}
\begin{figure*}[h]
\centering
\begin{tikzpicture}
\def\qualTopGap{-2pt}
\def\qualSideGap{2pt}

\node[anchor=south west, inner sep=0] (img)
    at (0.05\textwidth,0)
    {\includegraphics[width=0.95\textwidth]
    {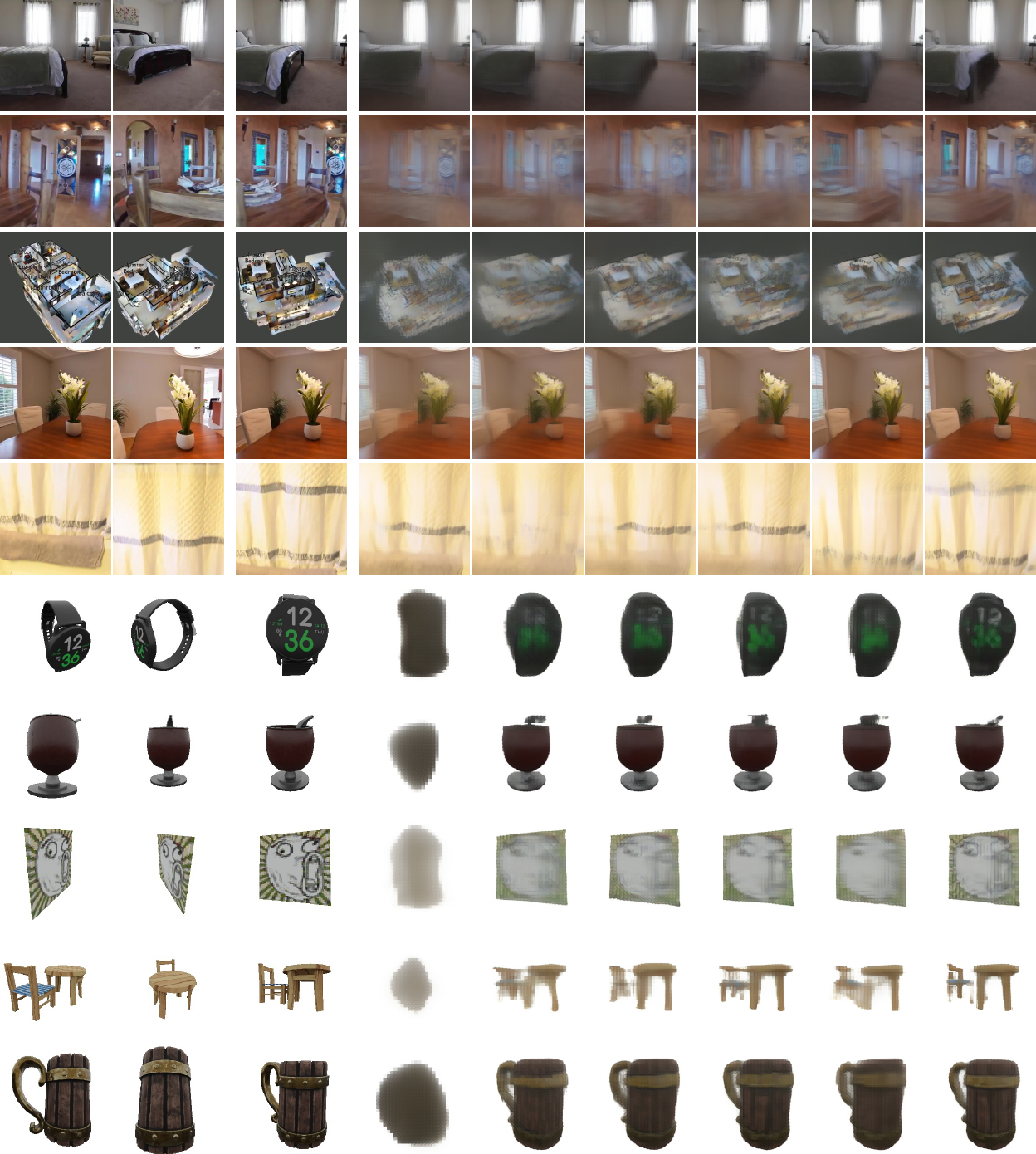}};

\foreach \x/\name in {
    0.10810/{Reference},
    0.28130/{GT},
    0.39983/{Pl\"ucker},
    0.50918/{GTA},
    0.61853/{PRoPE},
    0.72788/{RayRoPE},
    0.83723/{URoPE},
    0.94658/{\ourmethod{}}
} {
    \node[anchor=south, font=\scriptsize\bfseries]
        at ($(img.north west)!\x!(img.north east)
            +(0,\qualTopGap)$)
        {\strut\name};
}

\node[rotate=90, anchor=south, font=\footnotesize\bfseries]
    at ($(img.south west)!0.75112!(img.north west)
        +(-\qualSideGap,0)$)
    {RE10K};

\node[rotate=90, anchor=south, font=\footnotesize\bfseries]
    at ($(img.south west)!0.24888!(img.north west)
        +(-\qualSideGap,0)$)
    {Objaverse};

\end{tikzpicture}
\caption{
Qualitative comparison of novel-view synthesis on RE10K and Objaverse.
The top five rows show RE10K examples, and the bottom five rows show
Objaverse examples. The first two columns contain the reference views,
followed by the target ground truth and predictions from the compared methods.
}
\label{fig:lvsm_main_qualitative}
\end{figure*}

Figure~\ref{fig:lvsm_main_qualitative} presents additional novel-view synthesis results on RE10K~\citep{zhou2018stereo} and Objaverse~\citep{deitke2023objaverse}.

\clearpage

\begin{figure*}[h]
\centering
\begin{tikzpicture}
\def\qualSubGap{-2pt}     
\def\qualGroupGap{6pt}    
\def\qualSingleGap{2pt}   
\def\qualSideGap{2pt}

\node[anchor=south west, inner sep=0] (img)
    at (0.05\textwidth,0)
    {\includegraphics[width=0.95\textwidth]
    {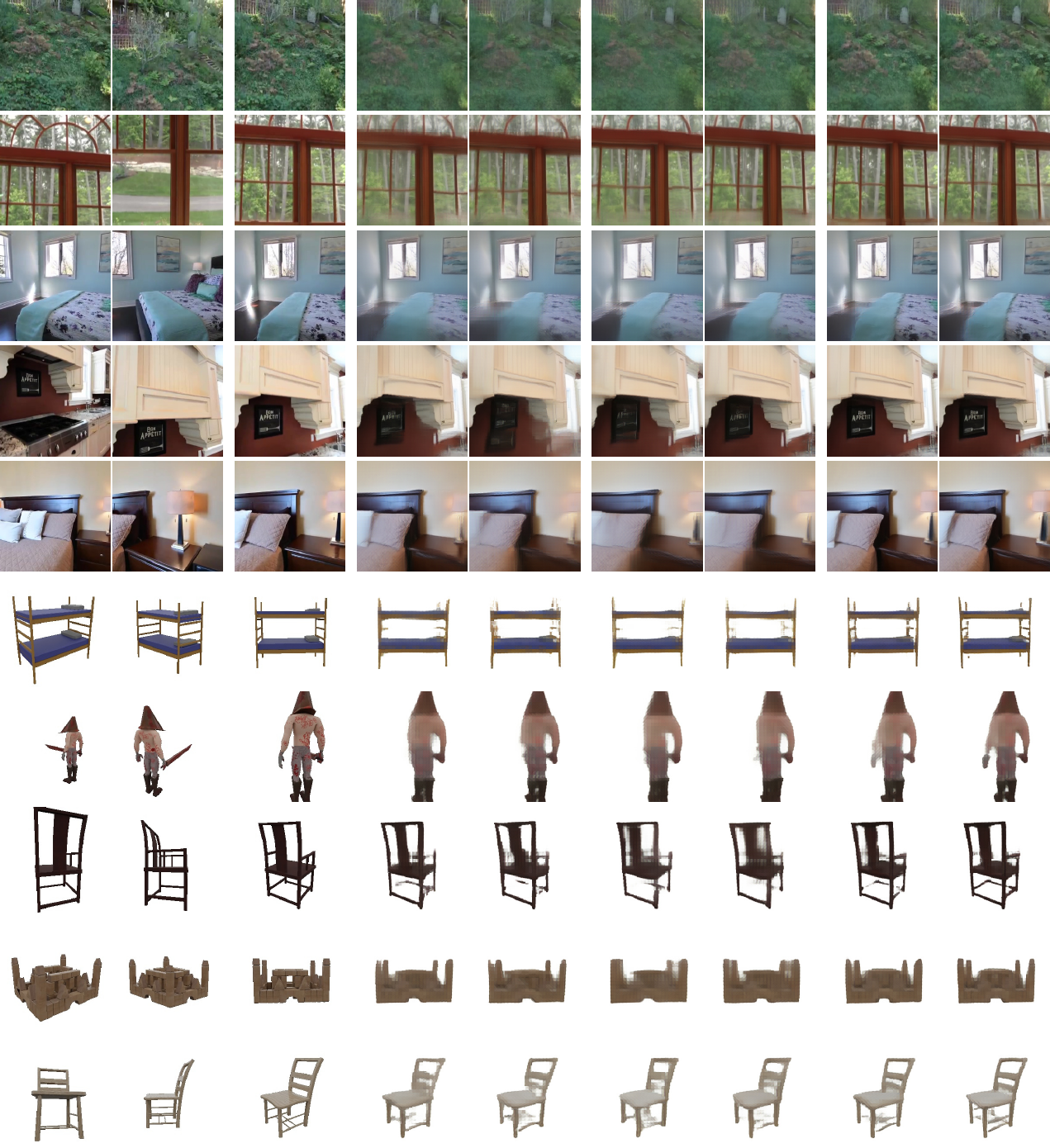}};

\foreach \x/\name in {
    0.10615/{Reference},
    0.27623/{GT}
} {
    \node[anchor=south, font=\scriptsize\bfseries]
        at ($(img.north west)!\x!(img.north east)
            +(0,\qualSingleGap)$)
        {\strut\name};
}

\foreach \x/\name in {
    0.44631/{RayRoPE},
    0.67008/{URoPE},
    0.89385/{\ourmethod{} (ours)}
} {
    \node[anchor=south, font=\scriptsize\bfseries]
        at ($(img.north west)!\x!(img.north east)
            +(0,\qualGroupGap)$)
        {\strut\name};
}

\foreach \x/\name in {
    0.39262/{+ Axial},
    0.50000/{+ Mixed},
    0.61639/{+ Axial},
    0.72377/{+ Mixed},
    0.84016/{+ Axial},
    0.94754/{+ Mixed}
} {
    \node[anchor=south, font=\scriptsize\bfseries]
        at ($(img.north west)!\x!(img.north east)
            +(0,\qualSubGap)$)
        {\strut\name};
}

\node[rotate=90, anchor=south, font=\footnotesize\bfseries]
    at ($(img.south west)!0.75112!(img.north west)
        +(-\qualSideGap,0)$)
    {RE10K};

\node[rotate=90, anchor=south, font=\footnotesize\bfseries]
    at ($(img.south west)!0.24888!(img.north west)
        +(-\qualSideGap,0)$)
    {Objaverse};

\end{tikzpicture}
\caption{
Qualitative comparison of the axial and mixed RoPE variants on RE10K
and Objaverse. The top five rows show RE10K examples, and the bottom five
rows show Objaverse examples. For each positional encoding, the axial and mixed RoPE variants are shown side by side.
}
\label{fig:lvsm_mixed_qualitative}
\end{figure*}

Figure~\ref{fig:lvsm_mixed_qualitative} presents qualitative novel-view synthesis results for the axial and mixed RoPE~\citep{heo2024ropevit} variants on RE10K~\citep{zhou2018stereo} and Objaverse~\citep{deitke2023objaverse}.

\clearpage

\subsection{Depth estimation}
\label{sec:unimatch_qual}

\begin{figure*}[h]
\centering
\begin{tikzpicture}
\def\qualTopGap{2pt}
\def\qualSideGap{2pt}

\node[anchor=south west, inner sep=0] (img)
    at (0.05\textwidth, 0)
    {\includegraphics[width=0.90\textwidth]{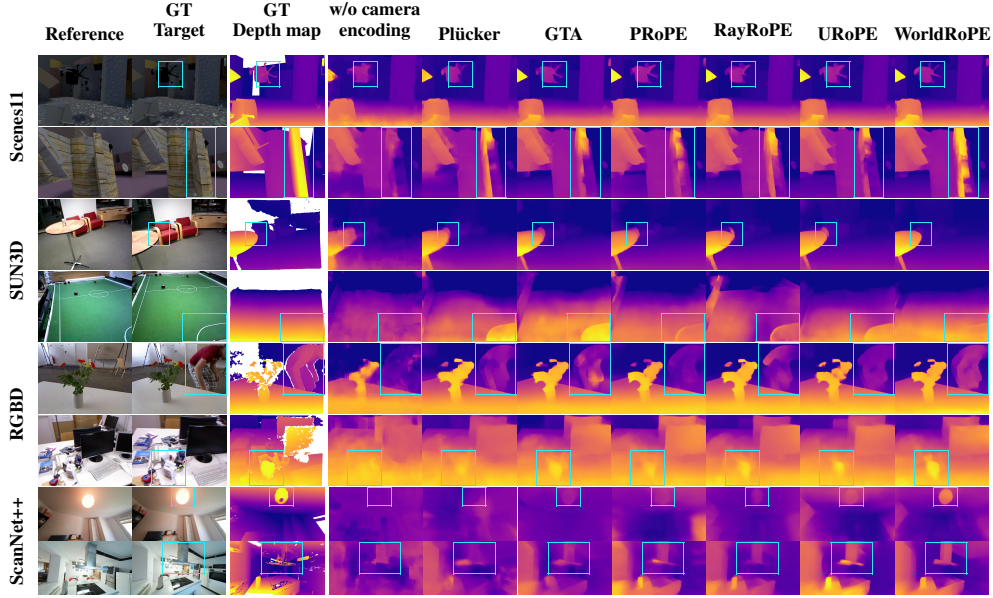}};

\foreach \x/\name in {
    0.049246/{Reference},
    0.148661/{\shortstack{GT\\Target}},
    0.251462/{\shortstack{GT\\Depth map}},
    0.354263/{\shortstack{w/o camera\\encoding}},
    0.453678/{\shortstack{Pl\"ucker}},
    0.553093/{GTA},
    0.652508/{PRoPE},
    0.751924/{RayRoPE},
    0.851339/{URoPE},
    0.950754/{\ourmethod{}}
} {
    \node[anchor=south, font=\scriptsize\bfseries]
        at ($(img.north west)!\x!(img.north east)+(0,\qualTopGap)$)
        {\name};
}

\foreach \y/\dataset in {
    0.868221/{Scenes11},
    0.601410/{SUN3D},
    0.334599/{RGBD},
    0.099783/{ScanNet++}
} {
    \node[rotate=90, anchor=south, font=\scriptsize\bfseries]
        at ($(img.south west)!\y!(img.north west)
            +(-\qualSideGap,0)$)
        {\dataset};
}

\path[use as bounding box] (0,0)
    rectangle ($(img.north west)+(0.95\textwidth,18pt)$);
\end{tikzpicture}
\caption{
Qualitative comparison of depth estimation on Scenes11, SUN3D, RGBD,
and unseen ScanNet++. The row groups follow this dataset order from
top to bottom. The first three columns contain the reference image,
target image, and ground-truth depth, followed by predictions from
the compared methods.}
\label{fig:unimatch_depth_qualitative}
\end{figure*}

Figure~\ref{fig:unimatch_depth_qualitative} presents qualitative depth estimation results on Scenes11~\citep{ummenhofer2017demon}, SUN3D~\citep{xiao2013sun3d}, RGBD~\citep{sturm2012benchmark}, and unseen ScanNet++~\citep{yeshwanth2023scannet++}.

\end{document}